\documentclass{article}

\usepackage[table,dvipsnames]{xcolor}
\usepackage[final]{corl_2026}

\usepackage{amsmath,amsfonts,amssymb}
\usepackage{booktabs}
\usepackage{multirow}
\usepackage{makecell}
\usepackage{array}
\usepackage{graphicx}
\usepackage{subcaption}
\usepackage{wrapfig}
\usepackage{pifont}
\usepackage{tikz}
\usepackage{lineno}
\usepackage{algorithm}
\usepackage{algpseudocode}
\usepackage{textcomp}
\usepackage{url}
\usepackage{verbatim}
\usepackage{xspace}
\usepackage{mathtools}
\usepackage{colortbl}
\usepackage{fontawesome5}

\definecolor{localfocus}{RGB}{220,238,255}
\definecolor{poolingbase}{RGB}{232,232,232}
\definecolor{oursrow}{RGB}{220,245,220}

\newcommand{\methodname}{FocusPool}

\usepackage{ifthen}
\newboolean{reftoapp}
\setboolean{reftoapp}{true}
\newboolean{reftomain}
\setboolean{reftomain}{true}

\title{Localized Visual Feature Aggregation via Focus Pooling for Visuomotor Policies}
\author{
Ruiyu Wang$^{*}$ \quad
Zheyu Zhuang \quad
Danica Kragic \quad
Florian T. Pokorny
\\
KTH Royal Institute of Technology, Sweden
\\
$^{*}$\texttt{ruiyuw@kth.se}
}

\begin{document}
\maketitle


\begin{abstract}
    Focusing on spatially localized, control-relevant visual cues has been shown to improve data efficiency in visuomotor policies by reducing the need to model task-irrelevant visual variation.
Existing methods often impose this focus through input preprocessing, such as cropping control- or object-centric regions in RGB images or point-clouds.
However, it remains underexplored whether such localized features can be exposed \textit{directly} from commonly used convolutional neural network (CNN) encoded features.
In this paper, we show that intermediate CNN features preserve localized visual context for control, but existing pooling methods fail to aggregate it effectively.
We introduce \methodname{}, an attention pooling module that selectively aggregates intermediate visual features according to their relevance to the robot's current proprioceptive context.
The resulting pooled representation captures task-progressive, control-relevant local information and is used directly for policy learning.
Across simulation and real-world experiments, \methodname{} improves policy success rates over pooling and explicit local focus methods by 36.2\% and 41.2\%, with training only 5.8\% of encoder parameters.
\href{https://github.com/RuiyuWANG/FocusPool}{\faGithub\ GitHub}
\end{abstract}


\section{Introduction}
Convolutional neural networks (CNNs) are commonly used as visual encoders in visuomotor policies, transforming high-dimensional image observations into spatial feature maps that are subsequently aggregated into compact representations for control~\cite{mandlekar_2021_robomimic,chi2023diffusionpolicy}.
This spatial compression poses a challenge in manipulation: small object- or contact-level cues may determine the next action, yet can be washed out when pooled with background, distractors, or objects irrelevant to the current subtask. 
Prior work addresses such dilution through coarse-to-fine designs that preprocess the input by cropping control-relevant regions from RGB images~\cite{james2022q,wang2026palm} or point clouds~\cite{james2022coarse,goyal2024rvt,hu2025generalizablecoarsetofinerobotmanipulation}, improving data efficiency.
However, such two-stage designs require additional training stages and can suffer from error accumulation. This motivates us to study end-to-end trainable pooling that extracts local spatial information directly from standard CNN feature maps.

\begin{figure}[t]
    \centering
    \includegraphics[width=0.8\linewidth]{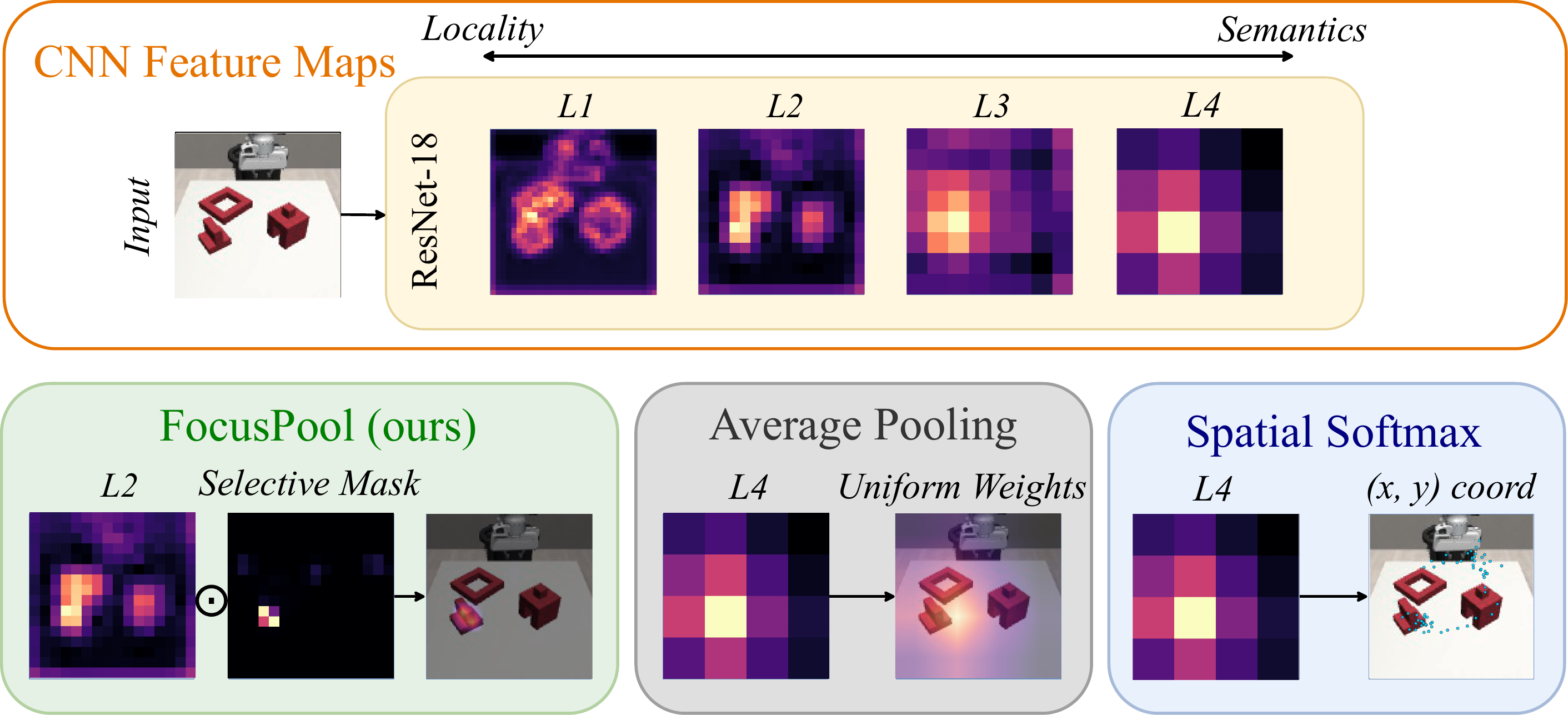}
    \caption{\textbf{Feature locality and pooling.}
    As convolution and downsampling accumulate, CNN feature maps become coarser and less spatially localized.
    \methodname{} selectively aggregates localized intermediate features, producing a progressive mask over localized task-focused regions.
    }
    \vspace{-3mm}
    \label{fig:teaser}
\end{figure}

A key challenge for end-to-end pooling, compared with explicit cropping, is the non-locality of CNN features. 
Concretely, deeper layers aggregate broader context and richer semantics but become less spatially localized~\cite{araujo2019computing}. 
This limits deep features for spatially precise prediction, motivating the use of intermediate, high-resolution features~\cite{hariharan2015hypercolumns,lin2017feature,wang2020deep}.
For visuomotor policies, applying existing pooling methods to intermediate features does not directly improve data efficiency, as demonstrated by our experiments in Fig.~\ref{fig:pooling-stage-success}. Although features are more localized, average pooling is not spatially selective~\cite{lin2013network}, while spatial softmax can introduce noisy keypoints~\cite{finn2016deep}, see Fig.~\ref{fig:teaser}.
Cross-attention provides a natural mechanism for selective aggregation. However, existing methods such as RcP~\cite{zhuang2024raising} and AFA~\cite{tsagkas2025attentive} do not produce localized task-progressive pooling masks.

To fill this gap, we introduce \methodname{}, a trainable attention-pooling module that aggregates localized, task-relevant regions from intermediate CNN features into compact representations. \methodname{} constructs a learnable robot-state-conditioned query and iteratively refines it through cross-attention with spatial visual patches. The resulting attention weights form a selective pooling mask over progressive, control-relevant regions (Fig.~\ref{fig:teaser}). 
By preserving control-relevant information while suppressing irrelevant spatial content, the compact representation improves policy data efficiency.
\methodname{} requires no input preprocessing, integrates with CNN encoders and different policy heads, and substantially reduces the number of trainable encoder parameters by pooling directly from an intermediate feature map, using only $5.8\%$ parameters of ResNet-18~\cite{he2016deep} in our main setting.

We evaluate \methodname{} on MimicGen~\cite{mandlekar2023mimicgen} and a real UFactory xArm7 robot. Our results show that:
(1) spatially localized intermediate CNN features can yield data-efficient policy representations, but existing pooling methods fail to exploit them effectively;
(2) the state-conditioned, iteratively refined cross-attention module implicitly extracts control-relevant local information when trained end-to-end; and
(3) \methodname{} improves policy success by 36.2\% in simulation and 41.2\% in the real world over pooling and explicit local-focus baselines under the same data budget, matches baseline performance with half the data, and trains only 5.8\% of the encoder parameters.

\section{Related Work}
\textbf{Local focus for visuomotor policies.}
Prior works commonly enforce local visual focus through coarse-to-fine designs, where a control-relevant region is first predicted and then used to restrict the visual input.
Q-attention methods use CNN-based Q-functions to select and zoom into task-relevant locations for RGB or point-cloud observations~\cite{james2022q,james2022coarse}.
RVT-2 predicts key-frame poses identified by robot state changes from multi-view representations and then applies a downstream policy to point clouds cropped around the predicted keypoints~\cite{goyal2024rvt}.
\citet{hu2025generalizablecoarsetofinerobotmanipulation} use language alignment to predict subtask keypoints, facilitating keypoint-centered cropping for long-horizon manipulation.
PALM aligns third-person RGB observations around the end-effector through fixed-size crops to train local policies with better generalization~\cite{wang2026palm}.
In contrast, our method does not preprocess the input; it exposes localized information already present in intermediate CNN features through trainable pooling.

\textbf{Attentive feature aggregation in robotics.}
Attention mechanisms are widely used to assign non-uniform weights to visual features for feature aggregation or pooling.
Vision Transformers aggregate information across image patches through self-attention~\cite{dosovitskiy2020image}, or use learned queries to select task-relevant visual information~\cite{carion2020end,jaegle2021perceiver}.
In robotics, SoftMP~\cite{yuan2021softmp} learns a spatial attention map to pool over repeatable local feature locations for place recognition.
Recent methods also use attention to perform multimodal sensor fusion.
Sparsh-X applies attentive pooling to tactile embeddings from multiple signals, including images, audio, motion, and pressure~\cite{higuera2025tactile}.
While \citet{jangir2022look} adopt Transformer-based cross-view attention to fuse egocentric and third-person visual feature maps.
Robot-centric Pooling uses proprioceptive states and contrastive learning to guide image feature aggregation for self-other distinction and distractor robustness~\cite{zhuang2024raising}.
AFA pools the frozen pretrained visual representation with a learnable pooling module that attends to task-relevant visual cues for out-of-domain robustness~\cite{tsagkas2025attentive}.
While these methods demonstrate the value of selective feature aggregation, \methodname{} focuses on localized aggregation of intermediate CNN features.

\section{Methodology}
\begin{figure}[t]
    \centering
    \includegraphics[width=0.85\linewidth]{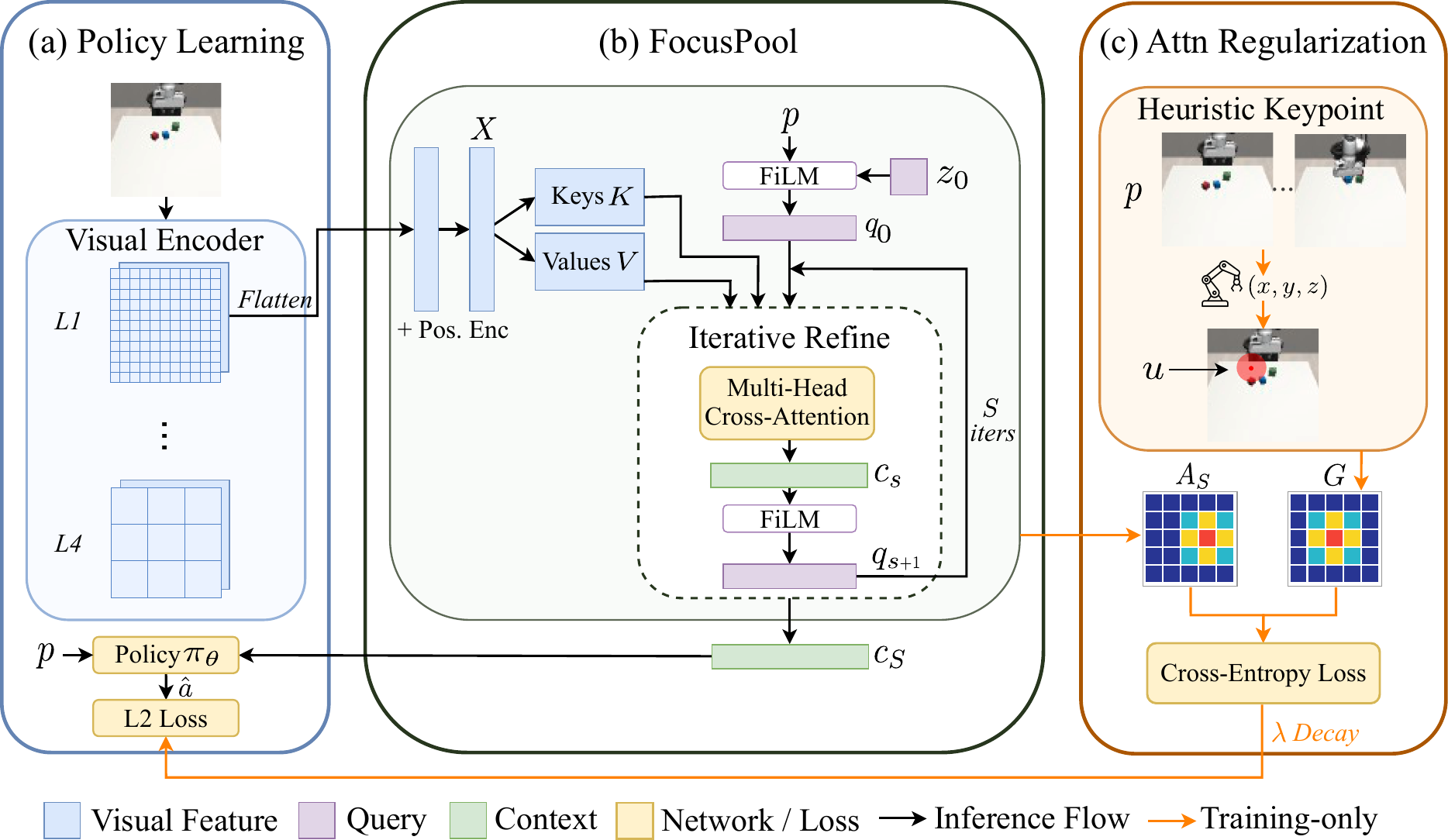}
    \caption{\textbf{\methodname{} overview.}
    (a) Intermediate visual features are leveraged for localized feature aggregation.
    (b) \methodname{} computes cross-attention between image-token keys/values and a robot-state-conditioned learnable query. The query is iteratively refined using the pooled context, and the final context is used as the policy input.
    (c) During early training, a weak regularizer biases the cross-attention weights toward heuristic control-relevant keypoints and quickly decays to zero.}
    \vspace{-3mm}
    \label{fig:method}
\end{figure}

\methodname{} is an attention-pooling module that produces compact, data-efficient control representations through localized feature aggregation, summarized in Fig.~\ref{fig:method}.
Below, we motivate intermediate layers as localized feature sources (\textit{where}), describe state-conditioned attention pooling for task-focused feature aggregation (\textit{how}), and introduce early attention regularization for training stability.

We consider end-to-end policy learning from demonstrations.
At each timestep $t$, the policy receives multi-view image observations $I_t=\{I_t^{m}\}_{m=1}^{M}$ and the robot proprioceptive state $p_t$, and predicts an action chunk $\hat{a}_t$.
Image observations are mapped to a compact visual representation $z_t=f_{\phi}(I_t)$ by the visual encoder $f_{\phi}$ and used by the policy to predict actions as $\hat{a}_{t}=\pi{\theta}(z_t,p_t)$.

\begin{wraptable}{r}{0.4\linewidth}
    \centering
    \vspace{-3mm}
    \resizebox{\linewidth}{!}{%
        \begin{tabular}{lcc}
\toprule
\textbf{Stage} & Resolution & Receptive Field \\
\midrule
Level 1 & $19\times19$ & $43\times43$  \\
Level 2 & $10\times10$ & $99\times99$  \\
Level 3 & $5\times5$   & $211\times211$  \\
Level 4 & $3\times3$   & $435\times435$ \\
\bottomrule
\end{tabular}
    }
    \vspace{-1mm}
    \caption{\textbf{ResNet-18 statistics.}
    Feature-map resolution and theoretical receptive field size~\cite{araujo2019computing} at different residual stages of ResNet-18 for $76\times76$ inputs.}
    \vspace{-5mm}
    \label{tab:resnet18-statistics}
\end{wraptable}

\subsection{\textit{Where to Aggregate}: Intermediate Visual Features for Locality}
\label{method:where}

A CNN-based visual encoder, e.g. ResNet~\cite{he2016deep}, produces a hierarchy of feature maps $F^{l} \in \mathbb{R}^{C_l \times H_l \times W_l}$ at feature levels $l \in {1,\ldots,L}$, where each level differs in spatial resolution and visual abstraction.
As residual blocks and downsampling stages accumulate, the output features have increasingly lower resolution and larger receptive fields, as shown in Tab.~\ref{tab:resnet18-statistics}.
Consequently, deeper feature maps aggregate increasingly non-local, scene-level information, which weakens their ability to expose localized features relevant to the immediate control signal.

We propose using intermediate features as the source for localized aggregation, as they retain more spatial structure while providing sufficient visual abstraction.
This design choice is further supported by experiments in Fig.~\ref{fig:pooling-stage-success}, where the gain of \methodname{} diminishes at deeper feature levels.

\subsection{\textit{How to Aggregate}: \methodname{}}
\label{method:how}

\methodname{} performs attention probing over spatial image patches using a learnable query modulated by robot state and iteratively injected visual context. State conditioning and iterative refinement provide the query with contextual information to assign higher weights to task-progressive, control-relevant patches, achieving implicit visual cropping via selective feature readout.

Specifically, a learnable latent token $z_0$ is randomly initialized and modulated by the robot state $p=[o^{\mathrm{eef}}, o^{\mathrm{grip}}]$ via FiLM transformation~\cite{perez2018film}, forming a robot-state-conditioned query $q_{0}$:
\[
q_{0}
\coloneqq
\mathrm{FiLM}(p,z_0)
=
(1+\gamma(p))\odot z_0+\beta(p),
\]
where $\gamma(\cdot)$ and $\beta(\cdot)$ are linear projection networks.

Intermediate feature map $F^{l}$ are flattened into visual features $\{f_n^{l}\}_{n=1}^{N=H_lW_l}$ each concatenated with a Fourier positional encoding $e_n$~\cite{tancik2020fourier} to preserve spatial information. The resulting image token is:
\[
x_n = [f_n^{l}; e_n],
\qquad
e_n \coloneqq
[\sin(\omega_b r_n),\cos(\omega_b r_n)]_{b=0}^{B-1},
\]
where $r_n \in [-1,1]^2$ is the normalized 2D coordinate of location $n$, and $\omega_b$ is the frequency at band $b$.
Stacking all tokens gives $X=[x_1,\ldots,x_N]^\top$.

A state-conditioned query $q_{0}$ can be ambiguous, as similar robot states may occur across different visual contexts or task stages, yielding diffuse focus when used for attention probing. 
\methodname{} progressively aligns the query with task progression by iteratively injecting visual context over $S$ refinement steps.
At each step $s$, the query $q_s$ and spatial image tokens $X$ are projected into queries, keys, and values, yielding head-wise attention and context:
\[
A_h
=
\mathrm{Softmax}\!\left(
d^{-1/2}Q_hK_h^\top
\right),
\qquad
C_h = A_hV_h.
\]
The query is then updated as:
\[
q_{s+1}
=
\mathrm{FiLM}\!\left(
\mathrm{LayerNorm}(c_s),\, q_s
\right),
\qquad s<S-1,
\]
where $c_s = \mathrm{Mean}_h(C_h)$ is the head-averaged context at step $s$.
After $S$ refinement steps, the final context $c_S$ serves directly as the visual representation for policy input.

\subsection{Early Attention Regularization}
\label{method:reg}

Early in training, a randomly initialized query may attend to arbitrary spatial tokens. 
Since intermediate CNN features retain broad receptive fields (Tab.~\ref{tab:resnet18-statistics}), these spurious locations can still provide predictive scene context and reinforce incorrect attention patterns.
Visual-focus convergence can therefore become sensitive to initialization, especially in tasks with limited spatial variation where policies may shortcut through robot state. 
We thus apply a weak attention prior during early training to stabilize mask convergence, whose contribution to the efficiency gains is analyzed in Tab.~\ref{tab:attention-reg}.

Following RVT-2~\cite{goyal2024rvt}, we use the next future keyframe end-effector position as a proxy control point and project it into each camera view.
For each valid view $m \in \mathcal{V}$, we construct a normalized Gaussian target heatmap $G^m$ on the pooling grid, centered at the projected target $u^{m,\mathrm{grid}}$:
\[
G^m(i,j)
\propto
\exp\left(
-\frac{\|(i,j)-u^{m,\mathrm{grid}}\|_2^2}{2\sigma^2}
\right),
\qquad
\sum_{i,j}G^m(i,j)=1 .
\]
Given the head-averaged attention map $\bar{A}^m$, the attention prior loss is:
\[
\mathcal{L}_{\mathrm{prior}}
=
-\frac{1}{|\mathcal{V}|}
\sum_{m\in\mathcal{V}}
\sum_{i,j}
G^m(i,j)
\log \left(\bar{A}^m(i,j)+\epsilon\right).
\]

We linearly decay $\lambda(k)$ from $\lambda_0$ to $0$, so the heuristic target has no further impact after $k$ warm-up steps and does not act as a persistent hand-designed prior.
The final training objective is:
\[
\mathcal{L}
=
\mathcal{L}_{\mathrm{policy}}
+
\lambda(k)\mathcal{L}_{\mathrm{prior}} .
\]

\section{Experiment}
\label{exp:sim}
\begin{table}[t]
    \centering
    \footnotesize
    \setlength{\tabcolsep}{1.9pt}
    \begin{tabular}{lccccccc}
\toprule
\textit{Method} & Thread. D2 & Square D2 & M. Cleanup D1 & Stack Three D1 & 3-Piece D2 & Pick Place D0$^\dagger$ & \textit{Average} \\
\midrule
\multicolumn{8}{c}{\textit{Explicit Local-focus Baseline}} \\
RVT-2~\cite{goyal2024rvt} & 12.7 & 26.0 & 46.0 & \underline{71.3} & \underline{26.7} & \underline{49.3} & \underline{38.7} \\
\midrule
\multicolumn{8}{c}{\textit{Pooling Baselines}} \\
AvgPool~\cite{lin2013network} & 20.7 & 22.0 & 46.7 & 52.7 & 14.0 & 29.0 & 30.9 \\
SSM~\cite{finn2016deep} & 24.0 & 26.7 & 49.3 & 66.7 & 17.3 & 30.0 & 35.7 \\
RcP~\cite{zhuang2024raising} & 11.3 & 13.3 & 36.0 & 38.7 & 2.0 & 5.3 & 17.8 \\
AFA~\cite{tsagkas2025attentive} & \underline{25.3} & \underline{28.7} & \underline{50.0} & 64.7 & 16.7 & 30.7 & 36.0 \\
\addlinespace[0.5pt]

\rowcolor{oursrow}
\methodname{} & \textbf{40.7} & \textbf{33.3} & \textbf{59.3} & \textbf{81.3} & \textbf{47.3} & \textbf{54.0} & \textbf{52.7} \\
\rowcolor{oursrow}
\textit{Gain (rel, \%)} & \textcolor{blue}{$60.5_{\uparrow}$} & \textcolor{blue}{$16.3_{\uparrow}$} & \textcolor{blue}{$18.7_{\uparrow}$} & \textcolor{blue}{$14.1_{\uparrow}$} & \textcolor{blue}{$77.3_{\uparrow}$} & \textcolor{blue}{$9.5_{\uparrow}$} & \textcolor{blue}{$36.2_{\uparrow}$} \\
\bottomrule
\end{tabular}
    \vspace{1mm}
    \caption{\textbf{Data efficiency on MimicGen.}
    Success rate (\%) and relative gain of \methodname{}-L2 over the best baseline (underlined). 
    Per-slot values are averaged best success rates across 3 training seeds.
    $^\dagger$Pick Place uses full-task completion success instead of the default accumulated subtask success.}
    \label{tab:exp-main-comparsion}
    \vspace{-4mm}
\end{table}

\textbf{Baselines.}
We compare \methodname{} against four feature-pooling baselines:
(1) average pooling (AvgPool)~\cite{lin2013network}, which globally averages spatial features;
(2) spatial softmax (SSM)~\cite{finn2016deep}, which maps convolutional activations to expected keypoint locations to preserve spatial information;
(3) robot-centric pooling (RcP)~\cite{zhuang2024raising} uses proprioception as a cross-attention query over image features and contrastive learning to pool the spatial region most aligned with the robot state;
and (4) attentive feature aggregation (AFA)~\cite{tsagkas2025attentive}, the most conceptually similar baseline, which pools frozen pretrained visual features via cross-attention with a learnable token.
All pooling baselines aggregate visual features from the latest CNN layer in their original design.
We also include RVT-2~\cite{goyal2024rvt} as a reference for explicit local focus through input preprocessing.
RVT-2 follows a coarse-to-fine design in which a first-stage multi-view transformer predicts an area of interest, after which zoomed-in views around this region are used for precise pose prediction.
We adapt this localization-and-cropping strategy to multi-view RGB observations, forming an RVT-2-style RGB baseline.
Details are in App.~\ref{app:implement_detail}.

\subsection{Simulation Evaluation}

\begin{wrapfigure}{r}{0.36\textwidth}
    \vspace{-0.5cm}
    \centering
    \includegraphics[width=\linewidth]{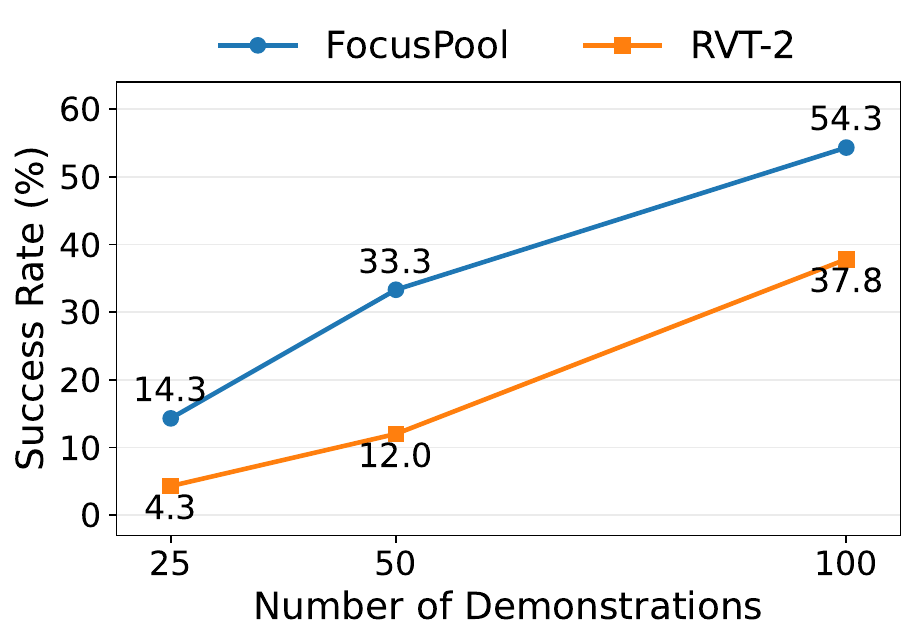}
    \caption{\textbf{Data efficiency curve.} Six-task average success rate of \methodname{} and RVT-2 using seed 0.}
    \label{fig:data_efficiency}
    \vspace{-0.5cm}
\end{wrapfigure}

\textbf{Experimental setup.}
We evaluate on six MimicGen tasks~\cite{mandlekar2023mimicgen} involving long-horizon manipulation or high spatial variation, see Fig.~\ref{fig:sim_task}.
All methods use a CNN-based Diffusion Policy~\cite{chi2023diffusionpolicy} with an ImageNet-pretrained ResNet-18 encoder~\cite{deng2009imagenet}, differing only in the pooling module or input preprocessing.
Unless otherwise specified, \methodname{} pools L2 features, while pooling baselines use L4 features following their standard configurations.
All policies are trained on 100 demonstrations, except in Fig.~\ref{fig:data_efficiency}.
AFA's encoder is finetuned, since freezing it as in the original recipe resulted in \textit{near-zero} success on this benchmark.
We report the maximum rollout success rate over 500 training epochs, evaluating every 10 epochs across 50 environments.
Regularization loss weight for \methodname{} starts at $2\times10^{-4}$ and decays linearly to 0 over the $5000$ training steps.
Details are in App.~\ref{app:implement_detail}.

\textbf{Data efficiency.}
We benchmark \methodname{} on challenging MimicGen tasks with a 100-demo budget.
As shown in Tab.~\ref{tab:exp-main-comparsion}, \methodname{} achieves a 36.2\% average relative gain over the per-task best baseline.
Fig.~\ref{fig:data_efficiency} further confirms the data-efficiency trend, with \methodname{} reaching comparable baseline performance with half the data.
Both RVT-2 and \methodname{} outperform the four pooling baselines, indicating the benefit of localized visual focus.
Although \methodname{} uses RVT-2-style keypoints as a weak early prior, these proxies can be imperfect; e.g., in \textit{Threading}, the interaction region is offset from the end-effector, where RVT-2 achieves only 12.7\% success compared with 40.7\% for \methodname{}.
This suggests that \methodname{} can adaptively refine its focus beyond the heuristic prior, whereas explicit two-stage cropping can fall short.
Overall, feature-level localized aggregation improves data efficiency.

\begin{table}[t]
    \centering
    \resizebox{\linewidth}{!}{
\begin{tabular}{lccccccc}
\toprule
\textit{Method}
& Thread. 
& Square 
& M. Cleanup 
& Stack Three 
& 3-Piece 
& Pick Place 
& \textit{Average} \\
\midrule

FocusPool with Reg.
& \textbf{40.7} / \textbf{46}
& 33.3 / 36
& \textbf{59.3} / \textbf{64}
& \textbf{81.3} / 82
& \textbf{47.3} / 50
& \textbf{54.0} / \textbf{60}
& \textbf{52.7} / \textbf{56.3} \\

FocusPool no Reg.
& 29.3 / 34
& 37.3 / 38
& 46.7 / 52
& 80.0 / \textbf{84}
& 43.3 / 48
& 40.7 / \textbf{60}
& 46.2 / 52.7 \\

FocusPool noisy Reg.
& 37.3 / 42
& \textbf{42.0} / \textbf{48}
& 49.3 / 52
& 80.0 / \textbf{84}
& 46.0 / \textbf{52}
& 50.0 / 58
& 50.8 / 52.7 \\

AFA with Reg.
& 28.0 / 30
& 26.7 / 32
& 39.3 / 42
& 61.3 / 64
& 12.0 / 12
& 32.6 / 40
& 33.3 / 36.7 \\

AFA no Reg.
& 25.3 / 26
& 28.7 / 30
& 50.0 / 56
& 64.7 / 68
& 16.7 / 20
& 30.7 / 32
& 36.0 / 38.7 \\

\bottomrule
\end{tabular}
}
    \vspace{1mm}
    \caption{
    \textbf{The effect of early attention regularization.}
    Per-task averaged / best success rates (\%) over 3 seeds for the full \methodname{}, \methodname{} without early attention regularization, \methodname{} with $\pm5\,\mathrm{cm}$ noise added to each keyframe coordinate used for regularization ($16.7\%$ of the workspace), and AFA with and without \methodname{}'s early attention regularization.
    }
    \vspace{-4mm}
    \label{tab:attention-reg}
\end{table}
\begin{figure}[t]
    \centering
    \subfloat[Policy success rate across pooling stages and pooling methods.]{
        \includegraphics[
            width=0.6\linewidth,
            keepaspectratio
        ]{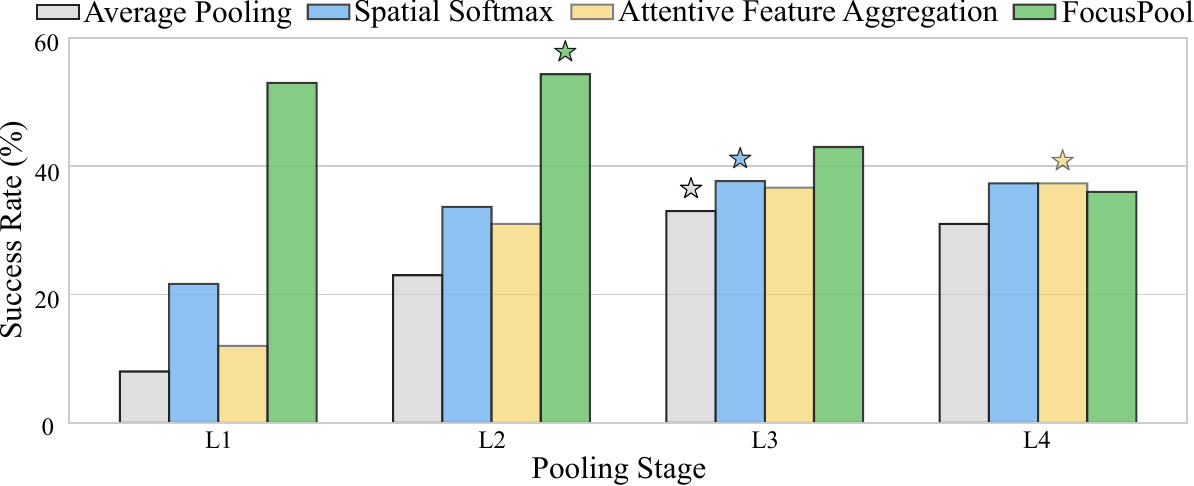}
        \label{fig:pooling-stage-success}
    }%
    \hfill%
    \subfloat[Feature activation at L2.]{
        \raisebox{1.2em}[0pt][0pt]{%
            \includegraphics[
                width=0.34\linewidth,
                keepaspectratio
            ]{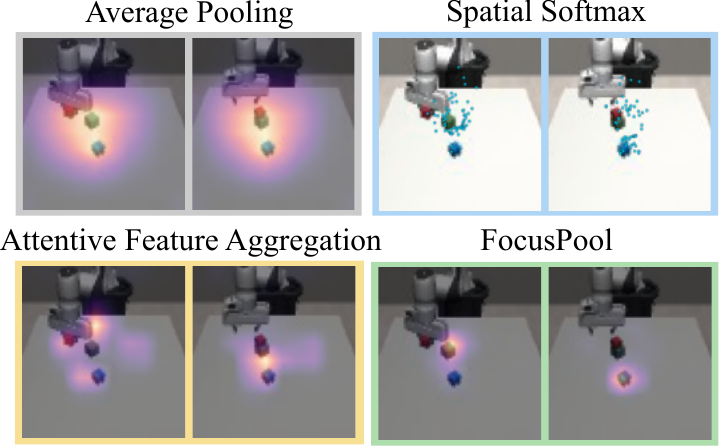}%
        }
        \label{fig:pooling-stage-vis}
    }
    \caption{\textbf{Pooling intermediate features.}
    (a) Policy success rates for four pooling methods when read out after different ResNet-18 residual stages, averaged over six tasks with seed 0; $\bigstar$ marks the best-performing stage for each method.
    Per-task result see Tab.~\ref{tab:seed0-pooling-stage}.
    (b) Pooling mask-weighted activations or spatial-softmax keypoints overlaid on the original input when read out from L2.
    }
    \vspace{-3mm}
    \label{fig:pooling-layers}
\end{figure}

\textbf{Are intermediate CNN features sufficient for localized pooling?}
To isolate the effect of using intermediate representations, we evaluate pooling baselines when read out from each CNN stage, as summarized in Fig.~\ref{fig:pooling-stage-success}.
All baselines perform best at later stages, whereas \methodname{} peaks at L1--L2.
Fig.~\ref{fig:pooling-stage-vis} provides qualitative insight into why only \methodname{} benefits from intermediate features.
At L2, \methodname{} concentrates on localized, control-relevant regions that shift with task progression, while the baselines produce diffuse attention or noisy keypoints and degrade when applied to less aggregated early-stage features.
This effect is more pronounced at L1, where \methodname{} achieves 52.0\% average success while training only 1.5\% of the encoder parameters, whereas the baselines drop substantially.
These results show that intermediate features alone are insufficient, and gains depend on selectively aggregating the localized information they preserve.

\begin{figure*}[h]
    \centering
    \subfloat[Success-rate degradation.]{
        \includegraphics[width=0.3\textwidth]{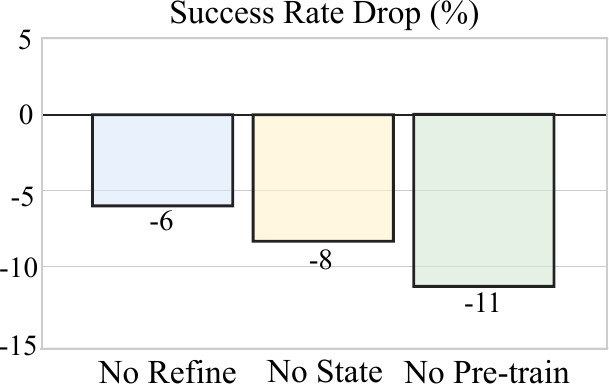}
        \label{fig:ablation-drop}
    }
    \hfill
    \subfloat[\methodname{}-L2 attention masks under ablated settings.]{
        \includegraphics[width=0.65\textwidth]{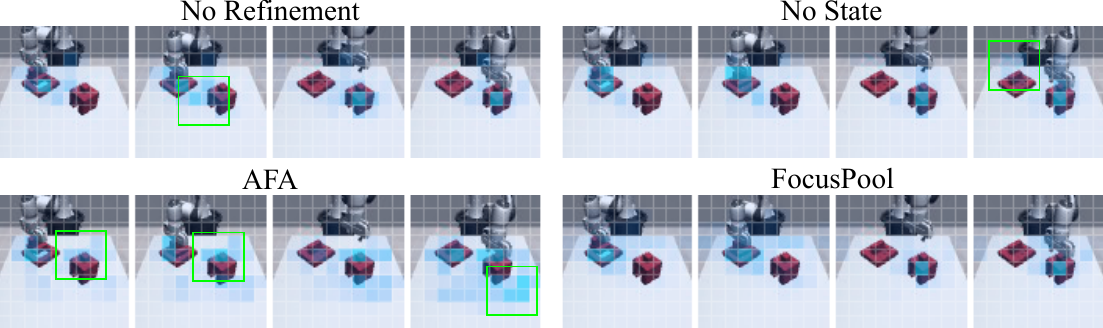}
        \label{fig:ablation-viz}
    }
    \caption{\textbf{Ablation study.}
    (a) Relative success-rate degradation when removing iterative refinement, robot-state conditioning, or pretrained encoder initialization from \methodname{}.
    (b) Qualitative changes in the learned \methodname{}-L2 attention masks under the corresponding ablations.
    }
    \vspace{-3mm}
    \label{fig:ablations}
\end{figure*}

\paragraph{How does FocusPool work?}
The task-focused aggregation of \methodname{} is enabled by two core designs, state conditioning and iterative refinement.
Together, they inject low-dimensional and visual context into the query to disambiguate task stages and learn task-progressive visual focus.
As shown in Fig.~\ref{fig:ablations}, removing either component makes the attention mask less selective and reduces success by 6 or 8 percentage points.
AFA at L2 can be viewed as a variant without either component, resulting in more diffuse attention and a larger 17-point performance gap.
Further discussion is provided in App.~\ref{app:attention_pooling_weights}.
Early attention regularization introduces a spatial prior, which itself does not explain the gain.
Tab.~\ref{tab:attention-reg} shows that applying it to AFA reduces average success from 36.0\% to 33.3\%.
For \methodname{}, regularization mainly improves training stability, increasing mean success by 6.5 points but best success by only 3.6 points.
The benefit is larger on tasks with limited spatial variation, such as \textit{Threading}, where policies may shortcut through robot state and make visual-focus convergence more initialization-sensitive.
Adding substantial noise to the prior yields similar performance, 50.8\% versus 52.7\%, indicating that coarse spatial guidance is sufficient.
Removing pretrained encoder weights also reduces performance in Fig.~\ref{fig:ablation-drop}, which we discuss further in Sec.~\ref{sec:limitation}.

\subsection{Real Evaluation}
\begin{figure}[h]
    \centering
    \subfloat[Task overview.]{
        \includegraphics[width=0.37\linewidth]{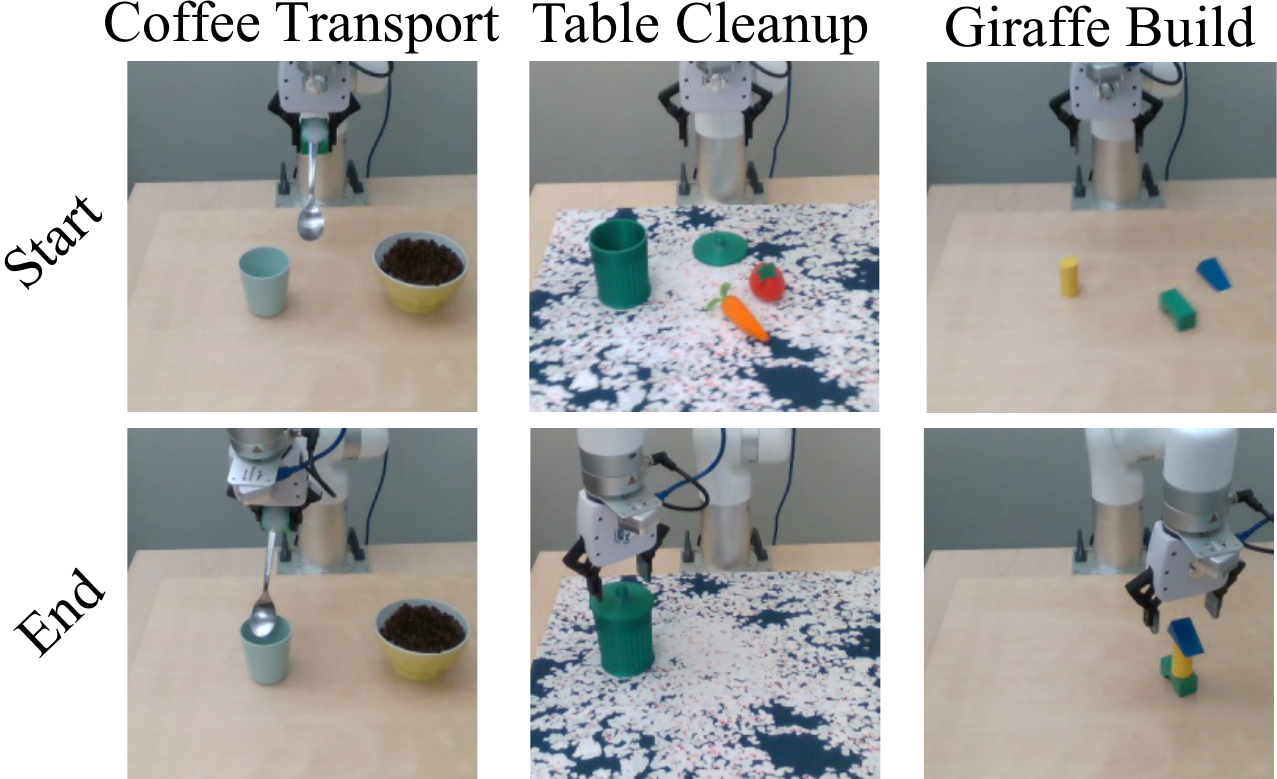}
        \label{fig:real-task-1}
    }
    \hfill
    \subfloat[\methodname{}-L2 attention masks.]{
        \includegraphics[width=0.32\linewidth]{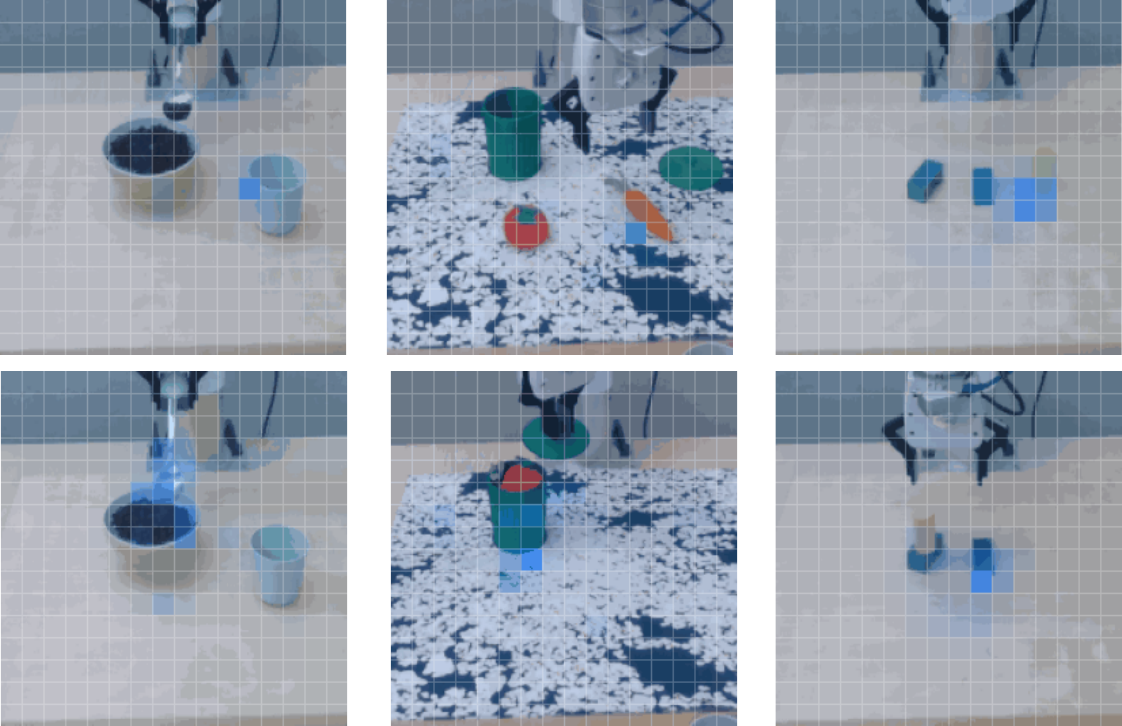}
        \label{fig:real-task-2}
    }
    \hfill
    \subfloat[Baseline failures.]{
        \includegraphics[width=0.21\linewidth]{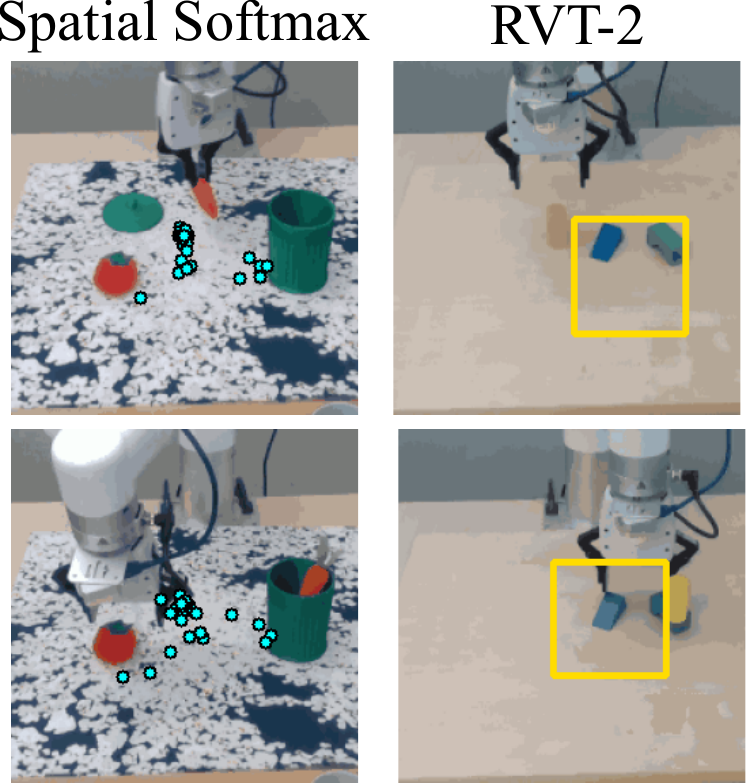}
        \label{fig:real-task-3}
    }
    \caption{\textbf{Real-world setup and evaluation.}
    (a) Example initial and completed states for three real-world tasks.
    (b) Learned \methodname{} attention-pooling masks when read out from L2.
    (c) Failure cases of spatial softmax and RVT-2, visualized with their predicted keypoints and zoomed-in areas.
    }
    \label{fig:real-tasks}
    \vspace{-3mm}
\end{figure}

\begin{table}[t]
    \centering
    \begin{tabular}{lccccc}
\toprule
\textit{Method} & \textit{Stage} & Coffee Transport & Table Cleanup & Giraffe Build & \textit{Average} \\
\midrule
RVT-2 & L4 & \underline{90.0} & \underline{60.0} & 20.0 & \underline{56.7} \\
Spatial Softmax  & L4 & 85.0 & 35.0 & \underline{35.0} & 51.7 \\
AFA (finetuned)  & L4 & \underline{90.0} & 30.0 & 30.0 & 50.0 \\
\rowcolor{oursrow}
\methodname{} & L1 & 95.0 & 70.0 & \textbf{70.0} & 78.3  \\
\rowcolor{oursrow}
\methodname{} & L2 & \textbf{100.0} & \textbf{75.0} & 65.0 & \textbf{80.0} \\
\rowcolor{oursrow}
\textit{Gain (rel, \%)} & -- & \textcolor{blue}{$11.1_{\uparrow}$} & \textcolor{blue}{$25.0_{\uparrow}$} & \textcolor{blue}{$100.0_{\uparrow}$} & \textcolor{blue}{$41.2_{\uparrow}$} \\
\bottomrule
\end{tabular}
    \vspace{1mm}
    \caption{\textbf{Real-world results.}
    Success rates (\%) and relative gains of \methodname{} over the best baseline (underlined).
    Each is evaluated over 20 real rollouts using the latest checkpoint (500 epochs).}
    \vspace{-4mm}
    \label{tab:real-exp}
\end{table}

\textbf{Experimental setup.}
We evaluate \methodname{} and three baselines in a multi-view RGB setup with a UFactory xArm~7, a fixed third-person camera, and an eye-in-hand camera.
As shown in Fig.~\ref{fig:real-task-1}, the designed tasks probe different aspects of localized visual feature aggregation.
\textit{Coffee Transport} tests continuous manipulation under offset target poses used for early attention regularization.
\textit{Table Cleanup} evaluates long-horizon manipulation in visually complex scenes.
\textit{Giraffe Build} stress-tests the proposed method on the high-precision small-object assembly task, where control-relevant visual cues are sparse and localized.
All models follow the same simulation training recipe.
We collect 100 demonstrations for \textit{Table Cleanup} and 50 for the other tasks.
Detailed descriptions are in App.~\ref{app:real-exp}.

\textbf{Real-world results.}
Tab.~\ref{tab:real-exp} shows that both \methodname{}-L1 and L2 outperform all baselines, achieving a 41.2\% average relative gain over the per-task best baseline, consistent with simulation.
Fig.~\ref{fig:real-task-2} further shows that the learned attention masks remain task-progressive and control-relevant in the real world.
In \textit{Coffee Transport}, although the regularization target is offset from the control-relevant spoon tip, \methodname{} corrects its focus during end-to-end training.
A similar advantage appears in \textit{Table Cleanup}, where visual complexity leads to noisy keypoint predictions for spatial softmax while \methodname{} maintains selective attention masks, as shown in Fig.~\ref{fig:real-task-2} and \ref{fig:real-task-3}.
The benefit is strongest on \textit{Giraffe Build}, where \methodname{} improves performance by 100\% and control-relevant cues occupy small image regions.
Moreover, L1 outperforms L2 only on this task, suggesting that finer spatial resolution and localised aggregation become particularly important when relevant visual signals are sparse.
In contrast, RVT-2 underperforms spatial softmax on \textit{Giraffe Build}, where keyframe-position errors can misplace the zoom-in region and cause the two-stage pipeline to fail.

\section{Discussion}
\textbf{Beyond ResNet-18 encoder.}
Our analysis is anchored on ResNet-18, a widely used CNN encoder for visuomotor policies.
Other CNN backbones, including deeper ResNets and architectures such as EfficientNet~\cite{tan2019efficientnet}, may exhibit different trade-offs between feature depth and spatial locality.
Our main comparison focuses on early and intermediate stages.
Since these stages preserve local spatial structure through similar convolution and pooling operations, the locality--semantic trade-off should broadly apply, at least for ResNet-based encoders.
Thus, our conclusion may extend to a broader class of CNN backbones: standard pooling tends to struggle with early or intermediate features, while localized task-relevant aggregation can better exploit them for efficient policy learning.

\textbf{General effectiveness of \methodname{}.}
The proposed cross-attention feature aggregation extends naturally to ViTs~\cite{dosovitskiy2020image} by attending the learned query over output patch tokens.
In Tab.~\ref{tab:exp_generality}, we evaluate a simple adaptation on the final-block patch tokens, where \methodname{} outperforms both the standard [CLS] representation and AFA.
The optimal readout stage, pretraining choice, and whether policies consume compact pooled features or full token sets may vary across architectures and are left for future study.
We further pair the pooled representation with a multimodal flow-matching policy head~\cite{lipman2022flow}, where similar gains persist.
Thus, the effectiveness of the proposed method extends across a broader range of encoders and policy heads.

\begin{table}[t]
    \centering
    \resizebox{\linewidth}{!}{
\begin{tabular}{llccccccc}
\toprule
\textit{Model} & \textit{Method}
& Thread.
& Square
& M. Cleanup
& Stack Three
& 3-Piece
& Pick Place
& \textit{Average} \\
\midrule

\multirow{3}{*}{\shortstack{ViT-S\\DINOv3}}
& [CLS]     & 18 & 26 & 30 & 46 & 14 & 46 & 30.0 \\
& AFA       & 12 & 18 & 36 & 46 & 14 & 50 & 29.3 \\
& FocusPool & 20 & 30 & 38 & 78 & 26 & 56 & 41.3 \\
\midrule

\multirow{3}{*}{\shortstack{Flow\\Matching}}
& AvgPool   & 28 & 26 & 46 & 64 & 14 & 26 & 34.0 \\
& AFA       & 24 & 22 & 42 & 52 & 12 & 30 & 30.7 \\
& FocusPool & 34 & 32 & 46 & 82 & 38 & 40 & 45.3 \\
\bottomrule

\end{tabular}
}
    \vspace{1mm}
    \caption{\textbf{Generality across encoder and policy head.}
    Success rates (\%) of AvgPool, AFA, and \methodname{} when applied to the output patch tokens of a DINOv3-pretrained ViT-S encoder, or when paired with a flow-matching policy head, using seed 0.
    For the ViT setting, \methodname{} is trained without attention regularization, and the [CLS] token serves as the standard ViT baseline.
    }
    \vspace{-4mm}
    \label{tab:exp_generality}
\end{table}

\section{Limitation}
\label{sec:limitation}

ImageNet pretraining clearly benefits \methodname{}, as shown in Fig.~\ref{fig:ablation-drop}.
Since it operates on intermediate features, pretrained weights provide useful low-level priors for early convolutional kernels, such as edge detectors learned from large-scale data, as also observed in prior works~\cite{wang2024equivariant,kleeraven}.
Even without pretraining, \methodname{} achieves $41.7\%$ success and outperforms the best baseline.
Pretraining also improves existing pooling methods, with a $40.5\%$ relative gain for diffusion policy with spatial softmax compared with $27.0\%$ for \methodname{}.
This indicates that \methodname{} is less dependent on, while still benefiting from pretrained local visual structure, detailed in App.~\ref{app:visual_pretrain}.
How different pretrained weights affect \methodname{} is left for future investigation.

\section{Conclusion}
We present \methodname{}, an end-to-end attention pooling module that extracts task-progressive, control-relevant local information from intermediate CNN feature maps.
Our analysis shows that such localized visual focus improves data efficiency and can emerge naturally through end-to-end training, without explicit input cropping induced by human-designed spatial priors.
Intermediate CNN features preserve useful spatial locality, but require selective readout to avoid diluting the less-aggregated spatial cues.
By integrating robot state and task progression through iteratively refined cross-attention probing, \methodname{} converts these early spatial features into more efficient control representations than deeper convolutional features aggregated without explicit control conditioning.

\section*{Acknowledgements}

This work was partially supported by the Wallenberg AI, Autonomous Systems and Software Program (WASP), funded by the Knut and Alice Wallenberg Foundation.

\bibliography{ref}

@inproceedings{tan2019efficientnet,
  title={Efficientnet: Rethinking model scaling for convolutional neural networks},
  author={Tan, Mingxing and Le, Quoc},
  booktitle={International Conference on Machine Learning (ICML)},
  year={2019},
}

@inproceedings{mandlekar_2021_robomimic,
  title={What Matters in Learning from Offline Human Demonstrations for Robot Manipulation},
  author={Mandlekar, Ajay and Xu, Danfei and Wong, Josiah and Nasiriany, Soroush and Wang, Chen and Kulkarni, Rohun and Fei-Fei, Li and Savarese, Silvio and Zhu, Yuke and Mart{\'\i}n-Mart{\'\i}n, Roberto},
  booktitle={Conference on Robot Learning (CoRL)},
year={2021}
}

@inproceedings{chi2023diffusionpolicy,
	title={Diffusion Policy: Visuomotor Policy Learning via Action Diffusion},
	author={Chi, Cheng and Feng, Siyuan and Du, Yilun and Xu, Zhenjia and Cousineau, Eric and Burchfiel, Benjamin and Song, Shuran},
	booktitle={Robotics: Science and Systems (RSS)},
	year={2023}
}

@article{mandlekar2023mimicgen,
  title={Mimicgen: A data generation system for scalable robot learning using human demonstrations},
  author={Mandlekar, Ajay and Nasiriany, Soroush and Wen, Bowen and Akinola, Iretiayo and Narang, Yashraj and Fan, Linxi and Zhu, Yuke and Fox, Dieter},
  journal={arXiv:2310.17596},
  year={2023}
}

@inproceedings{wang2024equivariant,
  author       = {Dian Wang and Stephen Hart and David Surovik and Tarik Kelestemur and Haojie Huang and Haibo Zhao and Mark Yeatman and Jiuguang Wang and Robin Walters and Robert Platt},
  title        = {Equivariant Diffusion Policy},
  booktitle    = {Conference on Robot Learning (CoRL)},
  year         = {2024},
}

@article{hu2025generalizablecoarsetofinerobotmanipulation,
      title={Generalizable Coarse-to-Fine Robot Manipulation via Language-Aligned 3D Keypoints},
      author={Jianshu Hu and Lidi Wang and Shujia Li and Yunpeng Jiang and Xiao Li and Paul Weng and Yutong Ban},
      year={2025},
      eprint={2509.23575},
      archivePrefix={arXiv},
      primaryClass={cs.RO},
}

@article{goyal2024rvt,
  author    = {Goyal, Ankit and Blukis, Valts and Xu, Jie and Guo, Yijie and Chao, Yu-Wei and Fox, Dieter},
  title     = {RVT2: Learning Precise Manipulation from Few Demonstrations},
  journal   = {Robotics: Science and Systems (RSS)},
  year      = {2024},
}

@inproceedings{kleeraven,
  title={RAVEN: End-to-end Equivariant Robot Learning with RGB Cameras},
  author={Klee, David and Hu, Boce and Cole, Andrew and Tian, Heng and Wang, Dian and Platt, Robert and Walters, Robin},
  booktitle={International Conference on Learning Representations (ICRL)},
  year={2026}
}

@inproceedings{he2016deep,
  title={Deep residual learning for image recognition},
  author={He, Kaiming and Zhang, Xiangyu and Ren, Shaoqing and Sun, Jian},
  booktitle={Conference on Computer Vision and Pattern Recognition (CVPR)},
  year={2016}
}

@article{tsagkas2025attentive,
  title={Attentive Feature Aggregation or: How Policies Learn to Stop Worrying about Robustness and Attend to Task-Relevant Visual Cues},
  author={Tsagkas, Nikolaos and Sochopoulos, Andreas and Danier, Duolikun and Vijayakumar, Sethu and Kouris, Alexandros and Mac Aodha, Oisin and Lu, Chris Xiaoxuan},
  journal={arXiv preprint arXiv:2511.10762},
  year={2025}
}

@inproceedings{finn2016deep,
  title={Deep spatial autoencoders for visuomotor learning},
  author={Finn, Chelsea and Tan, Xin Yu and Duan, Yan and Darrell, Trevor and Levine, Sergey and Abbeel, Pieter},
  booktitle={International Conference on Robotics and Automation (ICRA)},
  year={2016},
}

@article{lin2013network,
  title={Network in network},
  author={Lin, Min and Chen, Qiang and Yan, Shuicheng},
  journal={arXiv preprint arXiv:1312.4400},
  year={2013}
}

@inproceedings{deng2009imagenet,
  title={Imagenet: A large-scale hierarchical image database},
  author={Deng, Jia and Dong, Wei and Socher, Richard and Li, Li-Jia and Li, Kai and Fei-Fei, Li},
  booktitle={Conference on Computer Vision and Pattern Recognition (CVPR)},
  year={2009},
}

@article{araujo2019computing,
  title={Computing receptive fields of convolutional neural networks},
  author={Araujo, Andr{\'e} and Norris, Wade and Sim, Jack},
  journal={Distill},
  year={2019}
}

@inproceedings{lin2017feature,
  title={Feature pyramid networks for object detection},
  author={Lin, Tsung-Yi and Doll{\'a}r, Piotr and Girshick, Ross and He, Kaiming and Hariharan, Bharath and Belongie, Serge},
  booktitle={Conference on Computer Vision and Pattern Recognition (CVPR)},
  year={2017}
}

@article{wang2020deep,
  title={Deep high-resolution representation learning for visual recognition},
  author={Wang, Jingdong and Sun, Ke and Cheng, Tianheng and Jiang, Borui and Deng, Chaorui and Zhao, Yang and Liu, Dong and Mu, Yadong and Tan, Mingkui and Wang, Xinggang and others},
  journal={IEEE Transactions on Pattern Analysis and Machine Intelligence},
  year={2020},
}

@inproceedings{hariharan2015hypercolumns,
  title={Hypercolumns for object segmentation and fine-grained localization},
  author={Hariharan, Bharath and Arbel{\'a}ez, Pablo and Girshick, Ross and Malik, Jitendra},
  booktitle={IEEE Conference on Computer Vision and Pattern Recognition (CVPR)},
  pages={447--456},
  year={2015}
}

@article{wang2026palm,
  author       = {Ruiyu Wang and
                  Zheyu Zhuang and
                  Danica Kragic and
                  Florian T. Pokorny},
  title        = {{PALM:} Enhanced Generalizability for Local Visuomotor Policies via
                  Perception Alignment},
  journal      = {IEEE Robotics and Automation Letters (R-AL)},
  year         = {2026},
}

@article{dosovitskiy2020image,
  title={An image is worth 16x16 words: Transformers for image recognition at scale},
  author={Dosovitskiy, Alexey and Beyer, Lucas and Kolesnikov, Alexander and Weissenborn, Dirk and Zhai, Xiaohua and Unterthiner, Thomas and Dehghani, Mostafa and Minderer, Matthias and Heigold, Georg and Gelly, Sylvain and others},
  journal={arXiv preprint arXiv:2010.11929},
  year={2020}
}

@inproceedings{carion2020end,
  title={End-to-end object detection with transformers},
  author={Carion, Nicolas and Massa, Francisco and Synnaeve, Gabriel and Usunier, Nicolas and Kirillov, Alexander and Zagoruyko, Sergey},
  booktitle={European Conference on Computer Vision (ECCV)},
  year={2020},
}

@inproceedings{jaegle2021perceiver,
  title={Perceiver: General perception with iterative attention},
  author={Jaegle, Andrew and Gimeno, Felix and Brock, Andy and Vinyals, Oriol and Zisserman, Andrew and Carreira, Joao},
  booktitle={International Conference on Machine Learning (ICML)},
  year={2021},
}

@inproceedings{yuan2021softmp,
  title={SoftMP: Attentive feature pooling for joint local feature detection and description for place recognition in changing environments},
  author={Yuan, Fangming and Neubert, Peer and Schubert, Stefan and Protzel, Peter},
  booktitle={IEEE International Conference on Robotics and Automation (ICRA)},
  year={2021},
}

@inproceedings{higuera2025tactile,
  title={Tactile beyond pixels: Multisensory touch representations for robot manipulation},
  author={Higuera, Carolina and Sharma, Akash and Fan, Taosha and Bodduluri, Chaithanya Krishna and Boots, Byron and Kaess, Michael and Lambeta, Mike and Wu, Tingfan and Liu, Zixi and Hogan, Francois Robert and others},
  booktitle={Conference on Robot Learning (CoRL)},
  year={2025},
}

@article{jangir2022look,
  title={Look closer: Bridging egocentric and third-person views with transformers for robotic manipulation},
  author={Jangir, Rishabh and Hansen, Nicklas and Ghosal, Sambaran and Jain, Mohit and Wang, Xiaolong},
  journal={IEEE Robotics and Automation Letters (R-AL)},
  year={2022},
  publisher={IEEE}
}

@inproceedings{zhuang2024raising,
  title={Raising Body Ownership in End-to-End Visuomotor Policy Learning via Robot-Centric Pooling},
  author={Zhuang, Zheyu and Kyrki, Ville and Kragic, Danica},
  booktitle={IEEE/RSJ International Conference on Intelligent Robots and Systems (IROS)},
  year={2024},
}

@inproceedings{james2022coarse,
  title={Coarse-to-fine q-attention: Efficient learning for visual robotic manipulation via discretisation},
  author={James, Stephen and Wada, Kentaro and Laidlow, Tristan and Davison, Andrew J},
  booktitle={IEEE/CVF Conference on Computer Vision and Pattern Recognition (CVPR)},
  year={2022}
}

@article{james2022q,
  title={Q-attention: Enabling efficient learning for vision-based robotic manipulation},
  author={James, Stephen and Davison, Andrew J},
  journal={IEEE Robotics and Automation Letters (R-AL)},
  year={2022},
}

@inproceedings{perez2018film,
  title={Film: Visual reasoning with a general conditioning layer},
  author={Perez, Ethan and Strub, Florian and De Vries, Harm and Dumoulin, Vincent and Courville, Aaron},
  booktitle={Proceedings of the AAAI Conference on Artificial Intelligence},
  year={2018}
}

@inproceedings{tancik2020fourier,
  title={Fourier features let networks learn high frequency functions in low dimensional domains},
  author={Tancik, Matthew and Srinivasan, Pratul and Mildenhall, Ben and Fridovich-Keil, Sara and Raghavan, Nithin and Singhal, Utkarsh and Ramamoorthi, Ravi and Barron, Jonathan and Ng, Ren},
  booktitle={Advances in Neural Information Processing Systems (NeurIPS)},
  year={2020}
}

@article{simeoni2025dinov3,
  title={Dinov3},
  author={Sim{\'e}oni, Oriane and Vo, Huy V and Seitzer, Maximilian and Baldassarre, Federico and Oquab, Maxime and Jose, Cijo and Khalidov, Vasil and Szafraniec, Marc and Yi, Seungeun and Ramamonjisoa, Micha{\"e}l and others},
  journal={arXiv preprint arXiv:2508.10104},
  year={2025}
}

@article{lipman2022flow,
  title={Flow matching for generative modeling},
  author={Lipman, Yaron and Chen, Ricky TQ and Ben-Hamu, Heli and Nickel, Maximilian and Le, Matt},
  journal={arXiv preprint arXiv:2210.02747},
  year={2022}
}

@article{mazzaglia2024redundancy,
  title={Redundancy-aware action spaces for robot learning},
  author={Mazzaglia, Pietro and et al.},
  journal={IEEE Robotics and Automation Letters (R-AL)},
  year={2024},
}

\clearpage
\ifthenelse{\boolean{reftoapp}}{
    \appendix
    \begin{center}
        {\Large\bfseries Appendix}
    \end{center}
    \vspace{1em}

\appendix

\setcounter{section}{0}
\setcounter{subsection}{0}
\setcounter{equation}{0}
\setcounter{figure}{0}
\setcounter{table}{0}

\renewcommand{\thesection}{\Alph{section}}
\renewcommand{\thesubsection}{\thesection.\arabic{subsection}}

\renewcommand{\thetable}{A.\Roman{table}}
\renewcommand{\thefigure}{A.\arabic{figure}}

\renewcommand{\theHsection}{appendix.\Alph{section}}
\renewcommand{\theHsubsection}{appendix.\Alph{section}.\arabic{subsection}}
\renewcommand{\theHfigure}{appendix.figure.\arabic{figure}}
\renewcommand{\theHtable}{appendix.table.\Roman{table}}
\renewcommand{\theHequation}{appendix.equation.\arabic{equation}}

\normalsize

\section{Pseudocode}
\label{app:pseudocode}
Algorithm~\ref{alg:method_pseudocode} summarizes the core \methodname{} pooling module.
It replaces the original pooling layer after residual stage $l$ and outputs the policy representation $c_S$ and averaged attention map $A$, which is used by the early attention regularization in Algorithm~\ref{alg:attention_regularization}.

\begin{algorithm}[h]
\caption{\methodname{} Pooling}
\label{alg:method_pseudocode}
\small
\begin{algorithmic}[1]
\Require CNN feature map $F^{l}$, proprioception $p$, steps $S$, query tokens $z_0$
\Ensure Visual representation $c_S$, attention map $A$

\State $X\leftarrow \{[f_n^l;e_n]\}_{n=1}^{H_lW_l}$
\State $(\gamma_0,\beta_0)\leftarrow\mathrm{MLP}_{\mathrm{state}}(p)$; \quad $q_1\leftarrow(1+\gamma_0)\odot z_0+\beta_0$
\State $K\leftarrow\mathrm{LN}(X)W^K$; \quad $V\leftarrow XW^V$

\For{$s=1,\ldots,S$}
    \State $Q_s\leftarrow W^Q\mathrm{LN}(q_s)$
    \For{each head $h$ and query token $r$}
        \State $A_{s,h,r}\leftarrow
        \mathrm{Softmax}\!\left(Q_{s,h,r}K_h^\top/\sqrt{d_h}\right)$
        \State $C_{s,h,r}\leftarrow A_{s,h,r}V_h$
    \EndFor
    \State $c_s\leftarrow\mathrm{Mean}_{h,r}(C_{s,h,r})$
    \If{$s<S$}
        \State $(\gamma_s,\beta_s)\leftarrow\mathrm{MLP}_{\mathrm{refine}}(\mathrm{LN}(c_s))$
        \State $q_{s+1}\leftarrow(1+\gamma_s)\odot q_s+\beta_s$
    \EndIf
\EndFor

\State $A\leftarrow\mathrm{Mean}_{h,r}(A_{S,h,r})$
\State \Return $c_S,A$
\end{algorithmic}
\end{algorithm}

\begin{algorithm}[h]
\caption{Early Attention Regularization}
\label{alg:attention_regularization}
\small
\begin{algorithmic}[1]
\Require Attention map $A\in\mathbb{R}^{B\times H_l\times W_l}$, target points $u\in\mathbb{R}^{B\times2}$, image size $D$, Gaussian width $\sigma$
\Ensure Prior loss $\mathcal{L}_{\mathrm{prior}}$

\State $g\leftarrow\left(\frac{u^x}{D-1}(W_l-1),\frac{u^y}{D-1}(H_l-1)\right)$
\State $G\leftarrow\mathrm{Norm}_{i,j}\!\left[
\exp\!\left(-\frac{(j-g^x)^2+(i-g^y)^2}{2\sigma^2}\right)
\right]$
\State $P\leftarrow\mathrm{Norm}_{i,j}[A]$
\State $\mathcal{L}_{\mathrm{prior}}\leftarrow
-\frac{1}{B}\sum_{b=1}^{B}\sum_{i,j}G_b(i,j)\log(P_b(i,j)+\epsilon)$
\State \Return $\mathcal{L}_{\mathrm{prior}}$
\end{algorithmic}
\end{algorithm}
\section{Implementation}
\label{app:implement_detail}

\subsection{Simulated Tasks}
The simulation benchmark consists of six MimicGen tasks~\cite{mandlekar2023mimicgen} with high spatial variation, as shown in Fig.~\ref{fig:sim_task}.
Please refer to the original paper for detailed task descriptions.

\begin{figure}[ht]
    \centering
    \includegraphics[width=1.0\linewidth]{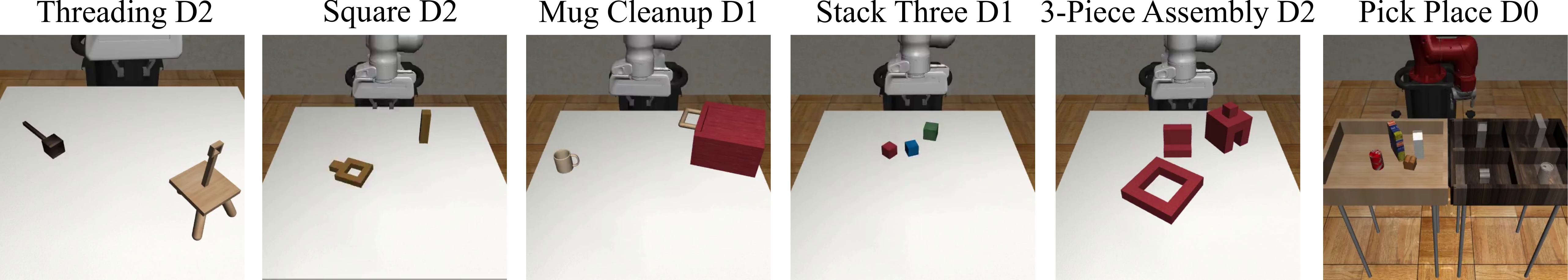}
    \caption{Overview of evaluated MimicGen tasks.}
    \label{fig:sim_task}
    \vspace{-3mm}
\end{figure}

\subsection{Real-world Setup}
\begin{figure}[t]
    \centering

    \begin{subfigure}{1.0\linewidth}
        \centering
        \includegraphics[width=1.0\linewidth]{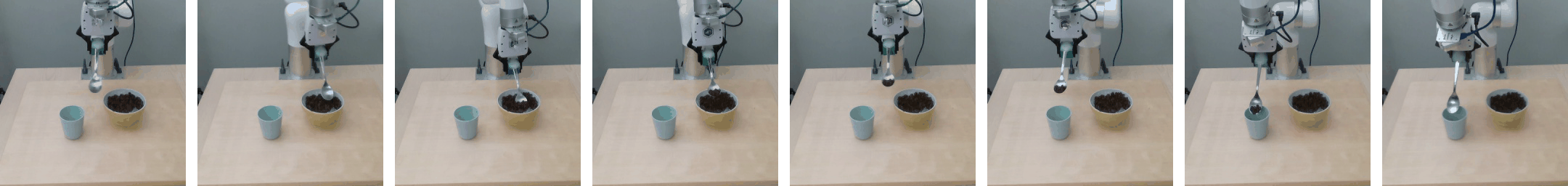}
        \caption{Coffee Transport}
        \label{fig:real-demo-coffee}
    \end{subfigure}

    \vspace{1mm}

    \begin{subfigure}{1.0\linewidth}
        \centering
        \includegraphics[width=1.0\linewidth]{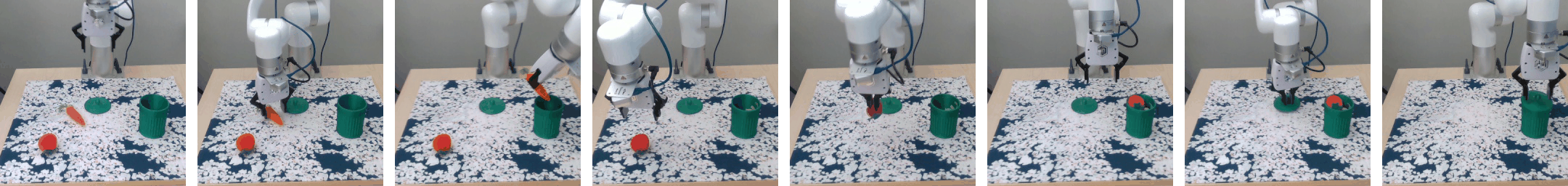}
        \caption{Table Cleanup}
        \label{fig:real-demo-cleanup}
    \end{subfigure}

    \vspace{1mm}

    \begin{subfigure}{1.0\linewidth}
        \centering
        \includegraphics[width=1.0\linewidth]{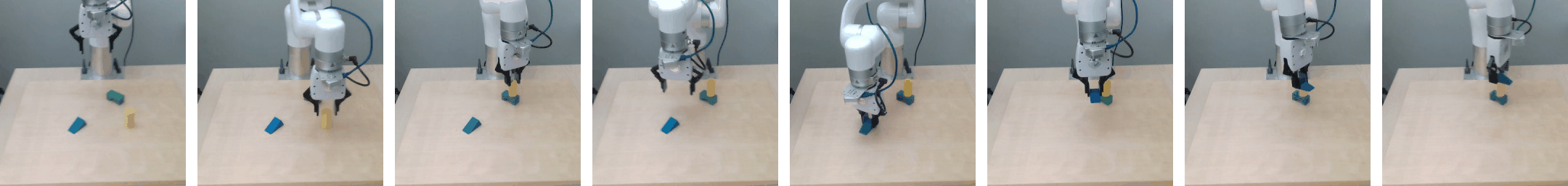}
        \caption{Giraffe Build}
        \label{fig:real-demo-giraffe}
    \end{subfigure}

    \caption{\textbf{Real-world task demonstrations.} The policies are trained on 50 demos for \textit{Coffee Transport} and \textit{Giraffe Build} and 100 demos for \textit{Table Cleanup}.}
    \label{fig:real-demo}
\end{figure}
We evaluate on three real-world tasks that involve different action modes and visual challenges.
A selected full demonstration of each task is shown in Fig.~\ref{fig:real-demo}.

\begin{itemize}
    \item \textit{Coffee Transport}: A continuous-motion task that requires scooping coffee beans from a bowl and pouring them into a small cup.
    \item \textit{Table Cleanup}: A long-horizon and visually complex task that requires picking and placing / inserting two toy vegetables into a bin and closing its lid.
    \item \textit{Giraffe Build}: A high-precision and small visual focus task that requires assembling three small blocks into a giraffe shape in four steps.
\end{itemize}

The tasks are executed on a UFactory xArm~7 under a multi-view RGB setup, with a fixed Intel RealSense D435i third-person camera and a RealSense D405 eye-in-hand camera.

\subsection{Training Detail}
We use an ImageNet-pretrained ResNet-18~\cite{he2016deep} as the visual encoder and Diffusion Policy~\cite{chi2023diffusionpolicy} as the policy head for the main evaluation.
Training follows the standard CNN-based diffusion policy setup, with the main policy training hyperparameters summarized in Tab.~\ref{tab:training_hyperparams}.

\begin{table}[h]
    \centering
    \begin{tabular}{ll}
    \toprule
    \textbf{Hyperparameter} & \textbf{Value} \\
    \midrule
    Training epoch & $500$ \\
    Batch size & $128$ \\
    Action representation & Absolute \\
    Observation steps & $2$ \\
    Action steps & $8$ \\
    Horizon & $16$ \\
    \bottomrule
    \end{tabular}
    \vspace{1mm}
    \caption{General hyperparameters for policy training.}
    \label{tab:training_hyperparams}
\end{table}

For both simulation and real-world experiments, the proprioceptive state consists of the end-effector 3D position, 6D rotation, and 1D gripper opening.
Actions are represented by the absolute end-effector position, 6D rotation, and gripper command in the workspace frame.
We do not evaluate joint-space actions, as prior work shows that end-effector action spaces provide better data efficiency and spatial generalization than joint-space control~\cite{mazzaglia2024redundancy}.

All methods use the same pooling strategy for both agent-view and eye-in-hand inputs.
For AvgPool, spatial softmax, RcP, AFA, and \methodname{}, simulated images are resized to $84\times84$ and randomly cropped to $76\times76$.
For real-world experiments, images are center-cropped to a square and resized to $128\times128$.
For RVT-2, simulation and real-world images are resized to $224\times224$, cropped with a $4\times$ zoom around the predicted keypoint with random center jitter, and then resized to the final policy input resolution after random cropping.

\subsection{Model Implementation}

\textbf{Spatial softmax.}
We implement spatial softmax following Robomimic~\cite{mandlekar_2021_robomimic} with $32$ keypoints.

\textbf{Robot-centric pooling.}
We reimplement RcP~\cite{zhuang2024raising} following its original image--proprioception alignment design.
The final ResNet-18 feature map is flattened into spatial tokens, while the proprioceptive state is encoded into a query that attends over these features.
RcP is trained with a MoCo-style contrastive objective to align image regions with robot state, using a momentum encoder and negative queue.
For policy pooling, the highest-scoring spatial location is selected as a one-hot mask and used to read out the corresponding visual feature, which is then concatenated with proprioception for downstream control.

\textbf{Attentive feature aggregation.}
AFA is an attention-pooling module proposed for out-of-domain generalization, where learned tokens cross-attend to the final feature map of frozen pretrained CNN or ViT encoders.
We adopt the official implementation of AFA~\cite{tsagkas2025attentive}.
We trained the AFA encoder end-to-end, since freezing the visual representation leads to near-zero performance on MimicGen. 

\textbf{RVT-2.}
RVT-2 uses trajectory-derived keyframes and local crops in its 3D manipulation pipeline~\cite{goyal2024rvt}.
We adapt this heuristic local-focus interface to RGB inputs, yielding an \emph{RVT-2-style cropping} baseline rather than a direct reimplementation of RVT-2.
The baseline trains a separate keypoint heatmap predictor before policy learning.
Given a $224\times224$ agent-view image and robot state, frozen DINOv3 patch features~\cite{simeoni2025dinov3} and robot-state tokens are fed to a Transformer that predicts a $14\times14$ local-focus heatmap.
The maximum-probability patch is then used as the fixed-size crop center for policy training and evaluation.
Following RVT-2, we use a zoom-in ratio of $4$, where the crop covers one quarter of the image width and height before being resized to the policy input resolution.

A keyframe is selected when the gripper state changes, when the robot joint velocity falls below a fixed threshold, or at the final frame.
We project the end-effector position at the selected future keyframe into the current agent-view image using the camera matrix.
The projected pixel location is mapped to the $14\times14$ heatmap grid and converted into a Gaussian-smoothed target.
The Transformer is trained with soft cross-entropy to predict this future end-effector heatmap.
We also use the same projected future-keyframe target as the heuristic keypoint for early attention regularization in \methodname{}.
The hyperparameters for the RVT-2-style cropping baseline are listed in Tab.~\ref{tab:rvt2_heatmap_hyperparams}.

\begin{table}[t]
    \centering
    \begin{tabular}{lll}
    \toprule
    \textbf{Category} & \textbf{Hyperparameter} & \textbf{Value} \\
    \midrule
    \multirow{6}{*}{Heatmap Model}
    & DINO image size & $224$ \\
    & Patch size & $16$ \\
    & Hidden dimension & $256$ \\
    & Transformer depth & $4$ \\
    & Transformer heads & $8$ \\
    & Transformer dropout & $0.1$ \\
    \midrule
    \multirow{3}{*}{Heuristic Keyframe}
    & Target Gaussian width & $12$ \\
    & Simulation joint-velocity threshold & $0.1$ \\
    & Real-world joint-velocity threshold & $0.025$ \\
    \bottomrule
    \end{tabular}
    \vspace{1mm}
    \caption{RVT-2 hyperparameters.}
    \label{tab:rvt2_heatmap_hyperparams}
\end{table}

\textbf{\methodname{}.}
\methodname{} is a state-conditioned attention pooling module applied to intermediate ResNet feature maps.
Its pooling query is conditioned on the 3D end-effector position and 1D gripper state.
At each forward pass, \methodname{} performs multi-head cross-attention between the query tokens and spatial image tokens, and averages the attended context over attention heads and query tokens to produce the policy representation.
The query is iteratively refined for $3$ steps using FiLM modulation from the pooled context.
The hyperparameters for \methodname{} are listed in Tab.~\ref{tab:focuspool_hyperparams}.

The early attention regularization uses the same projected future-keyframe target as RVT-2.
We project the end-effector position at the next heuristic keyframe into the current camera views and map the projected point to the pooling grid.
A Gaussian target heatmap is constructed around the projected point, and a cross-entropy loss is applied between this target heatmap and the head-averaged attention map produced by \methodname{}.
The regularization loss is added to the policy loss with a small initial weight, which linearly decays to zero over $5000$ trainging steps.

\begin{table}[t]
    \centering
    \begin{tabular}{llc}
        \toprule
        \textbf{Category} & \textbf{Hyperparameter} & \textbf{Value} \\
        \midrule
        \multirow{3}{*}{\methodname{}}
        & Refinement steps & $3$ \\
        & Attention heads & $4$ \\
        & Head dimension & $128$ \\
        \midrule
        \multirow{4}{*}{Attention Regularization}
        & Policy loss weight & $1.0$ \\
        & Initial regularization loss weight & $2.0\times10^{-4}$ \\
        & Loss decay steps & $5000$ \\
        & Gaussian width & $1.3$ \\
        \bottomrule
    \end{tabular}
    \vspace{1mm}
    \caption{\methodname{} hyperparameters.}
    \label{tab:focuspool_hyperparams}
\end{table}
\section{Further Discussion on Attention Pooling}
\label{app:ablation}

\subsection{Constructing Attention Pooling Masks}
\label{app:attention_pooling_weights}

Attentive Feature Aggregation (AFA) is the closest pooling baseline to \methodname{} in design, since it also uses cross-attention to aggregate image features.
However, AFA does not directly adaptively update the query with robot state and visual task-progression information.
When made trainable at an intermediate ResNet stage, AFA can be viewed as a simplified variant of \methodname{} without robot-state conditioning and query iterative refinement.
We further analyze how these two components affect the localization of aggregated visual features and, consequently, policy performance.
Fig.~\ref{fig:stack_three_ablation_viz} visualizes the attention pooling masks for the resulting variants of \methodname{}.

\begin{figure}[h]
    \centering
    \includegraphics[width=1.0\linewidth]{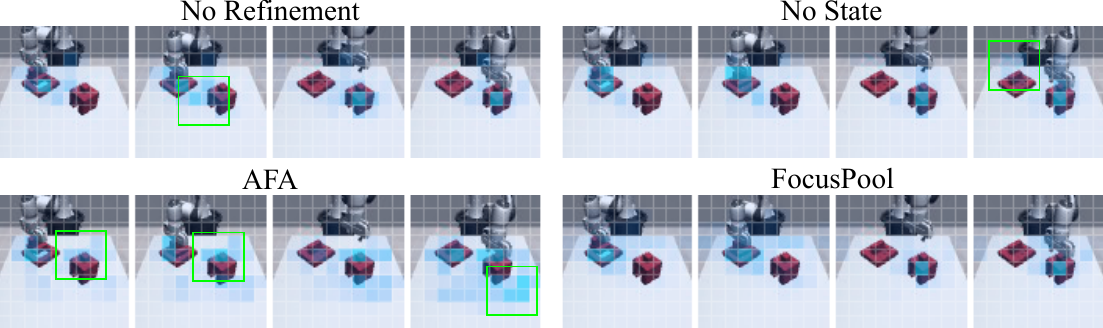}
    \caption{\textbf{Pooling mask visualization.}
    Attention pooling weights for \methodname{} variants on \textit{Stack Three} when read out after residual stage 2: no iterative refinement, no robot-state conditioning, baseline attentive feature aggregation, and the full \methodname{}.}
    \label{fig:stack_three_ablation_viz}
\end{figure}

Compared with the full \methodname{}, Fig.~\ref{fig:stack_three_ablation_viz} shows that removing iterative refinement yields a less progression-aware focus.
For example, in the highlighted green region, attention shifts partly to the lower block before the current insertion is completed.
This suggests that robot state alone can be ambiguous for identifying task progress in long-horizon manipulation.
With refinement, the pooled visual context is fed back into the query, allowing scene information to disambiguate the current stage.
Performance is comparable with 2 and 3 refinement steps in our sweep, so we use 3 steps to retain additional capacity for more demanding tasks at negligible cost.

The comparison with the no-state variant and AFA shows that iterative refinement can still improve the learned query without initial state conditioning.
However, removing state conditioning eliminates an important source of task-progress information, making attention more diffuse around subtask transitions.
In the highlighted region, the focus shifts to the next block before the gripper has closed.
This also causes a larger performance drop when task progress is difficult to infer from vision alone.
For example, on \textit{Stack Three}, the no-state variant drops by $16\%$, compared with $8\%$ for the no-iteration variant in Tab.~\ref{tab:focus-refine-ablation-seed0}.
The three blocks are visually similar but differ in height, making robot state a more explicit cue for the current task stage.

Removing both components makes AFA’s learned attention less localized and less task-progressive, leading to the largest drop in policy success.
The visualization also helps explain why AFA degrades when trained on intermediate features, supporting the view that the data-efficiency gain comes from localized feature readout.
Such task-progressive, control-relevant selection requires grounding from both robot state and visual context, which \methodname{} provides through state conditioning and iterative refinement.

\begin{table}[t]
    \centering
    \small
    \setlength{\tabcolsep}{2.8pt}
    \begin{tabular}{lcccccc}
    \toprule
    \textit{Method} & Mug Cleanup D1 & Square D2 & Threading D2 & Three Piece D2 & Stack Three D1 & Pick Place D0 \\
    \midrule
    No Iter. & 52 \textcolor{red}{(-4)} & \textbf{38} \textcolor{blue}{(+2)} & 34 \textcolor{red}{(-12)} & 42 \textcolor{red}{(-4)} & 74 \textcolor{red}{(-8)} & 50 \textcolor{red}{(-10)} \\
    No State & 52 \textcolor{red}{(-4)} & 36 (+0) & 28 \textcolor{red}{(-18)} & 40 \textcolor{red}{(-6)} & 66 \textcolor{red}{(-16)} & 54 \textcolor{red}{(-6)} \\
    AFA & 56 (+0) & 28 \textcolor{red}{(-8)} & 26 \textcolor{red}{(-20)} & 16 \textcolor{red}{(-30)} & 68 \textcolor{red}{(-14)} & 30 \textcolor{red}{(-30)} \\
    \methodname{} & \textbf{56} & 36 & \textbf{46} & \textbf{46} & \textbf{82} & \textbf{60} \\
    \bottomrule
    \end{tabular}
    \vspace{1mm}
    \caption{\textbf{Per-task ablation results} corresponding to Fig.~\ref{fig:ablation-drop}.
    Success rates (\%) for \methodname{} variants: no iterative refinement (no Iter.), no robot-state query conditioning (no State), baseline attentive feature aggregation (AFA), and the full \methodname{} on six MimicGen tasks using seed 0.}
    \vspace{-3mm}
    \label{tab:focus-refine-ablation-seed0}
\end{table}

\subsection{Effect of Visual Pre-training}
\label{app:visual_pretrain}
\begin{table}[h]
    \centering
    \small
    \setlength{\tabcolsep}{2.8pt}
    \begin{tabular}{lcccccc}
    \toprule
    \textit{Method} & Thread. & Square & M. Cleanup & Stack Three & 3-Piece & Pick Place \\
    \midrule
    SSM no pretarin & 17 & 8 & 43 & 38 & 4 & 20 \\
    SSM with pretrain & 24 & 27 & 49 & 67 & 17 & 30 \\
    \methodname{} no pretrain & 26 & \textbf{38} & 54 & 76 & 18 & 48  \\
    \methodname{} with pretrain & \textbf{46} & 36 & \textbf{56} & \textbf{82} & \textbf{46} & \textbf{60} \\
    \bottomrule
    \end{tabular}
    \vspace{2mm}
    \caption{\textbf{Pretraining effect.}
    Success rates (\%) for spatial softmax (SSM) and \methodname{} with and without encoder ImageNet pretraining weights on six MimicGen tasks using seed 0.}
    \vspace{-3mm}
    \label{tab:pretraining-ablation-seed0}
\end{table}

\methodname{} and all baselines are trained with pretrained encoders throughout the experiments.
Pretraining can improve data efficiency by providing useful low-level visual priors for early convolutional layers, such as edge detectors learned from large-scale data, as also observed in prior work~\cite{wang2024equivariant,kleeraven}.
We further analyze how pretraining affects the data efficiency of the spatial softmax baseline and \methodname{}, with results summarized in Tab.~\ref{tab:pretraining-ablation-seed0}.

Comparing each method with and without pretraining shows that pretraining improves performance for both spatial softmax and \methodname{}.
The gains are most pronounced on tasks with harder action modes for \methodname{}, such as \textit{Threading} and \textit{Three-Piece Assembly}.
spatial softmax, in contrast, shows consistent gains from pretraining across tasks.
On \textit{Square} and \textit{Stack Three}, \methodname{} performs similarly with and without pretraining, while pretraining improves spatial softmax performance by $19\%$ and $29\%$, respectively.
This suggests that \methodname{} is less dependent on pretrained features and can more efficiently aggregate task-relevant visual information from demonstrations.
\section{Full Simulation Result}
\label{app:sim_results}

\subsection{Data Efficiency}
\label{app:data_efficiency}

Tab.~\ref{tab:pooling-attn-prior-results} reports the results corresponding to Tab.~\ref{tab:exp-main-comparsion}, including standard deviations over three seeds.
Tab.~\ref{tab:per-task-data-efficiency} reports the per-task results for RVT-2 and \methodname{} trained on 25 and 50 demonstrations, corresponding to Fig.~\ref{fig:data_efficiency}.

\begin{table}[!h]
    \centering
    \small
    \setlength{\tabcolsep}{4pt}
    \begin{tabular}{lcccccc}
    \toprule
    \textit{Method} & Thread. D2 & Square D2 & M. Cleanup D1 & Stack Three D1 & 3P Assem. D2 & Pick Place D0\\
    \midrule
    \multicolumn{7}{c}{\textit{Local-Focus Baseline}} \\
    RVT2 & 12.7{\small$_{\pm 1.2}$} & 26.0{\small$_{\pm 2.0}$} & 46.0{\small$_{\pm 1.7}$} & \underline{71.3{\small$_{\pm 1.2}$}} & \underline{26.7{\small$_{\pm 1.2}$}} & \underline{49.3{\small$_{\pm 3.1}$}} \\
    \midrule
    \multicolumn{7}{c}{\textit{Pooling Baselines}} \\
    AvgPool & 20.7{\small$_{\pm 2.1}$} & 22.0{\small$_{\pm 3.5}$} & 46.7{\small$_{\pm 5.0}$} & 52.7{\small$_{\pm 5.4}$} & 14.0{\small$_{\pm 9.2}$} & 29.0{\small$_{\pm 5.3}$} \\
    SSM & 24.0{\small$_{\pm 3.5}$} & 26.7{\small$_{\pm 5.0}$} & 49.3{\small$_{\pm 1.2}$} & 66.7{\small$_{\pm 5.0}$} & 17.3{\small$_{\pm 1.2}$} & 30.0{\small$_{\pm 2.0}$}  \\
    AFA & \underline{25.3{\small$_{\pm 1.2}$}} & \underline{28.7{\small$_{\pm 1.2}$}} & \underline{50.0{\small$_{\pm 5.3}$}} & 64.7{\small$_{\pm 4.2}$} & 16.7{\small$_{\pm 3.1}$} & 30.7{\small$_{\pm 1.2}$} \\
    \rowcolor{oursrow}
    \methodname{} & \textbf{40.7{\small$_{\pm 3.2}$}} & \textbf{33.3{\small$_{\pm 1.8}$}} & \textbf{59.3{\small$_{\pm 4.8}$}} & \textbf{81.3{\small$_{\pm 2.2}$}} & \textbf{47.3{\small$_{\pm 6.7}$}} & \textbf{54.0{\small$_{\pm 2.2}$}} \\
    \bottomrule
    \end{tabular}
    \vspace{1mm}
    \caption{\textbf{Data efficiency on MimicGen.}
    Success rates (\%) of \methodname{} and baselines, reported as mean and standard deviation over three seeds using the best training-evaluation checkpoint.}
    \label{tab:pooling-attn-prior-results}
\end{table}

\begin{table}[!h]
    \centering
    \small
    \setlength{\tabcolsep}{4pt}
    \begin{tabular}{llcccccc}
    \toprule
    \textit{Model} & \textit{\# Demos}
    & Thread. 
    & Square 
    & M. Cleanup 
    & Stack Three 
    & 3P Assem. 
    & Pick Place  \\
    \midrule
    
    \multirow{2}{*}{RVT-2}
    & 25  & 8  & 2  & 8  & 6  & 0  & 2  \\
    & 50  & 8  & 6  & 26 & 14 & 6  & 12 \\
    \midrule
    \multirow{2}{*}{FocusPool}
    & 25  & 8  & 4  & 24 & 28 & 6  & 16 \\
    & 50  & 20 & 28 & 36 & 62 & 24 & 30 \\
    
    \bottomrule
    \end{tabular}
    \vspace{1mm}
    \caption{\textbf{Data efficiency curve.} Per-task success rate (\%) for RVT-2 and \methodname{} trained on 25 and 50 demons using seed 0.}
    \label{tab:per-task-data-efficiency}
\end{table}

\subsection{Pooling Stage}
Tab.~\ref{tab:seed0-pooling-stage} reports the pre-task results corresponding to Fig.~\ref{fig:pooling-stage-success}.

\begin{table*}[!h]
    \centering
    \normalsize
    \setlength{\tabcolsep}{3pt}
    \begin{subtable}{0.48\textwidth}
    \centering
    \label{tab:seed0-pooling-stage-avgpool}
    \begin{tabular}{lcccc}
    \toprule
    \textit{Task} & \textit{L1} & \textit{L2} & \textit{L3} & \textit{L4} \\
    \midrule
    Mug Cleanup D1 & 32 & 44 & \textbf{50} & 46 \\
    Square D2 & 4 & 22 & \textbf{32} & 22 \\
    Threading D2 & 8 & 22 & \textbf{28} & 20 \\
    Stack Three D1 & 2 & 32 & 48 & \textbf{54} \\
    Three Piece D2 & 0 & 2 & \textbf{18} & 14 \\
    Pick Place D0 & 2 & 16 & 22 & \textbf{30} \\
    \midrule
    \textit{Average} & 8 & 23 & \textbf{33} & 31 \\
    \bottomrule
    \end{tabular}
    \caption{Average pooling}
    \end{subtable}
    \hfill
    \begin{subtable}{0.48\textwidth}
    \centering
    \label{tab:seed0-pooling-stage-ssm}
    \begin{tabular}{lcccc}
    \toprule
    \textit{Task} & \textit{L1} & \textit{L2} & \textit{L3} & \textit{L4} \\
    \midrule
    Mug Cleanup D1 & 44 & \textbf{52} & \textbf{52} & 50 \\
    Square D2 & 22 & 24 & \textbf{32} & \textbf{32} \\
    Threading D2 & 20 & \textbf{30} & 28 & 22 \\
    Stack Three D1 & 36 & 56 & 62 & \textbf{72} \\
    Three Piece D2 & 4 & 10 & \textbf{24} & 18 \\
    Pick Place D0 & 4 & \textbf{30} & 28 & \textbf{30} \\
    \midrule
    \textit{Average} & 22 & 34 & \textbf{38} & 37 \\
    \bottomrule
    \end{tabular}
    \caption{Spatial softmax}
    \end{subtable}
    
    \vspace{1mm}
    
    \begin{subtable}{0.48\textwidth}
    \centering
    \label{tab:seed0-pooling-stage-afa}
    \begin{tabular}{lcccc}
    \toprule
    \textit{Task} & \textit{L1} & \textit{L2} & \textit{L3} & \textit{L4} \\
    \midrule
    Mug Cleanup D1 & 30 & 48 & 54 & \textbf{56} \\
    Square D2 & 10 & 26 & 24 & \textbf{28} \\
    Threading D2 & 16 & 24 & \textbf{28} & 26 \\
    Stack Three D1 & 10 & 46 & 62 & \textbf{68} \\
    Three Piece D2 & 0 & 16 & \textbf{26} & 16 \\
    Pick Place D0 & 6 & 26 & 26 & \textbf{30} \\
    \midrule
    \textit{Average} & 12 & 31 & \textbf{37} & \textbf{37} \\
    \bottomrule
    \end{tabular}
    \caption{Attentive feature aggregation}
    \end{subtable}
    \hfill
    \begin{subtable}{0.48\textwidth}
    \centering
    \label{tab:seed0-pooling-stage-focus-refine}
    \begin{tabular}{lcccc}
    \toprule
    \textit{Task} & \textit{L1} & \textit{L2} & \textit{L3} & \textit{L4} \\
    \midrule
    Mug Cleanup D1 & 48 & \textbf{56} & 54 & 52 \\
    Square D2 & \textbf{40} & 36 & 34 & 24 \\
    Threading D2 & 38 & \textbf{46} & 28 & 24 \\
    Stack Three D1 & \textbf{84} & 82 & 78 & 64 \\
    Three Piece D2 & \textbf{50} & 46 & 34 & 22 \\
    Pick Place D0 & 58 & \textbf{60} & 30 & 30 \\
    \midrule
    \textit{Average} & 53 & \textbf{54} & 43 & 36 \\
    \bottomrule
    \end{tabular}
    \caption{\methodname{}}
    \end{subtable}

    \caption{\textbf{Pooling intermediate features.} Policy success rates (\%) for four pooling methods when read out after different ResNet-18 residual stages, across six MimicGen tasks using seed 0.}
     \label{tab:seed0-pooling-stage}
\end{table*}
\section{Real-world Visualization}
\label{app:real-exp}

We visualize the learned spatial softmax keypoints, RVT-2 crop regions, and L2 \methodname{} attention-pooling weights during real-world policy rollouts on 3 tasks in Figs.~\ref{fig:real-coffee-attn}, \ref{fig:real-cleanup-attn}, and \ref{fig:real-giraffe-attn}.

\begin{figure}[t]
    \centering
    \begin{subfigure}{0.6\linewidth}
        \centering
        \includegraphics[width=1.0\linewidth]{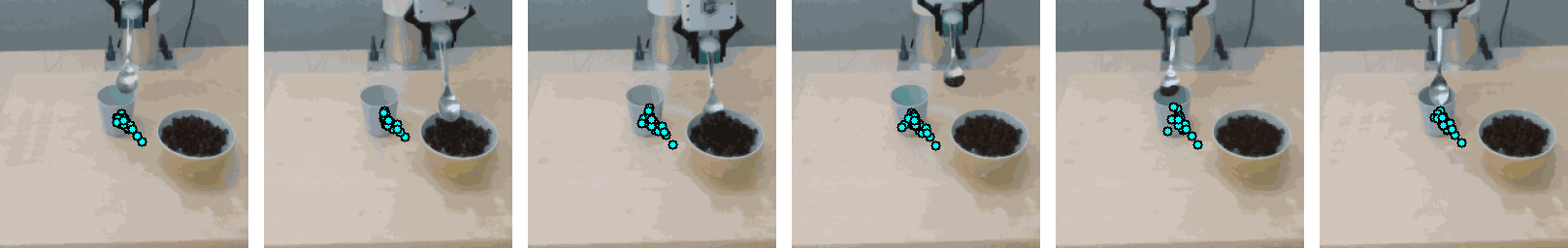}
        \caption{Spatial Softmax}
        \label{fig:coffee-spatial-softmax}
    \end{subfigure}

    \vspace{1mm}

    \begin{subfigure}{0.6\linewidth}
        \centering
        \includegraphics[width=1.0\linewidth]{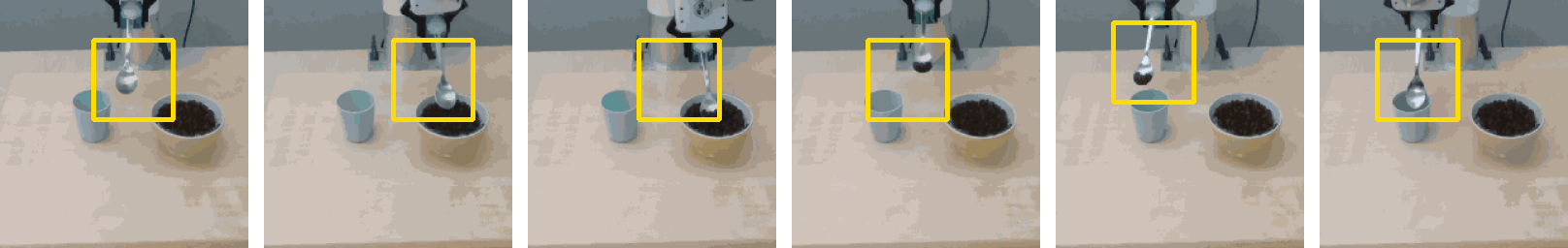}
        \caption{RVT-2}
        \label{fig:coffee-rvt2}
    \end{subfigure}

    \vspace{1mm}

    \begin{subfigure}{0.6\linewidth}
        \centering
        \includegraphics[width=1.0\linewidth]{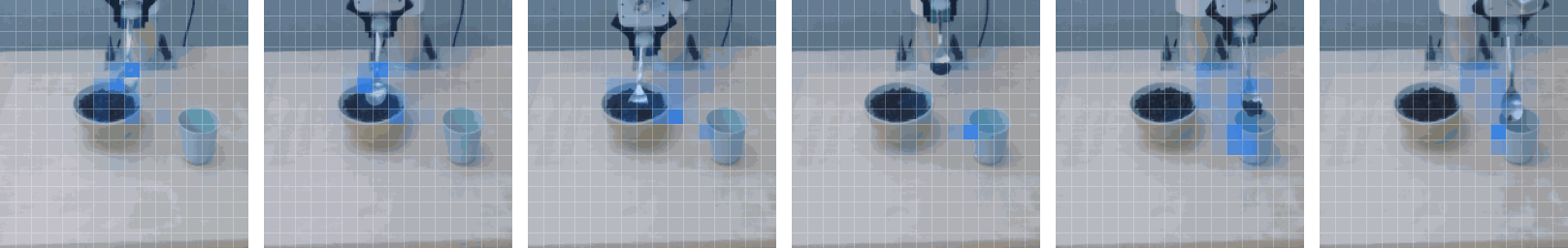}
        \caption{\methodname{}}
        \label{fig:coffee-focuspool}
    \end{subfigure}

    \caption{\textbf{Visual Focus on Coffee Transport.}
    Visualization of Spatial Softmax keypoints, RVT-2 predicted crop regions, and \methodname{} attention-pooling weights overlaid on the input images.}
    \label{fig:real-coffee-attn}
\end{figure}

\begin{figure}[h]
    \centering
    \begin{subfigure}{0.6\linewidth}
        \centering
        \includegraphics[width=1.0\linewidth]{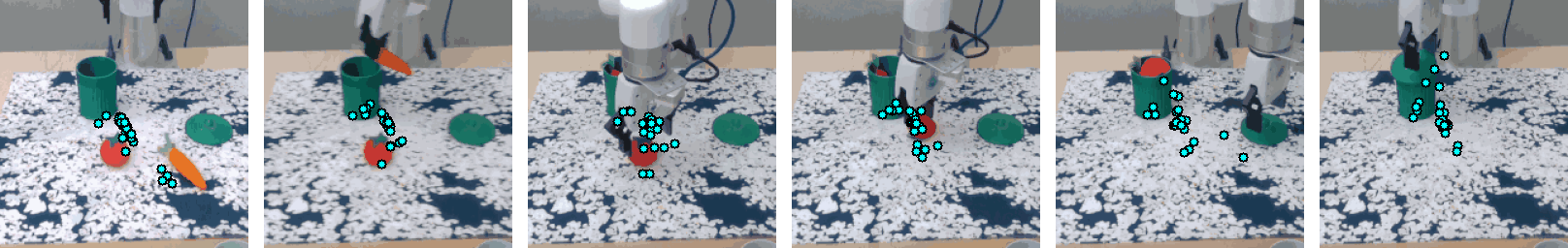}
        \caption{Spatial Softmax}
        \label{fig:cleanup-spatial-softmax}
    \end{subfigure}

    \vspace{1mm}

    \begin{subfigure}{0.6\linewidth}
        \centering
        \includegraphics[width=1.0\linewidth]{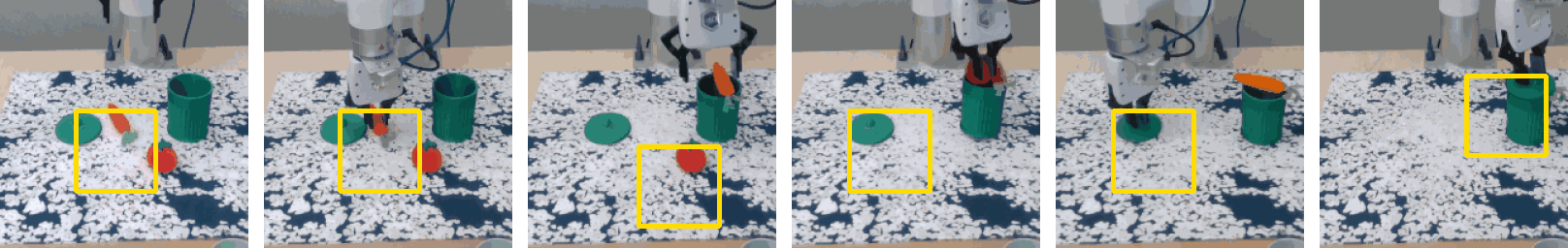}
        \caption{RVT-2}
        \label{fig:cleanup-rvt2}
    \end{subfigure}

    \vspace{1mm}

    \begin{subfigure}{0.6\linewidth}
        \centering
        \includegraphics[width=1.0\linewidth]{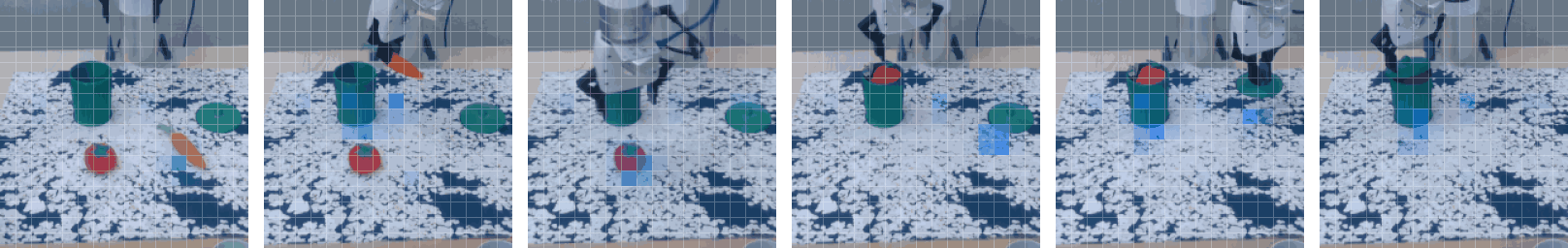}
        \caption{\methodname{}}
        \label{fig:cleanup-focuspool}
    \end{subfigure}

    \caption{\textbf{Visual Focus on Table Cleanup.}
    Visualization of Spatial Softmax keypoints, RVT-2 predicted crop regions, and \methodname{} attention-pooling weights overlaid on the input images.}
    \label{fig:real-cleanup-attn}
\end{figure}

\begin{figure}[h]
    \centering
    \begin{subfigure}{0.6\linewidth}
        \centering
        \includegraphics[width=1.0\linewidth]{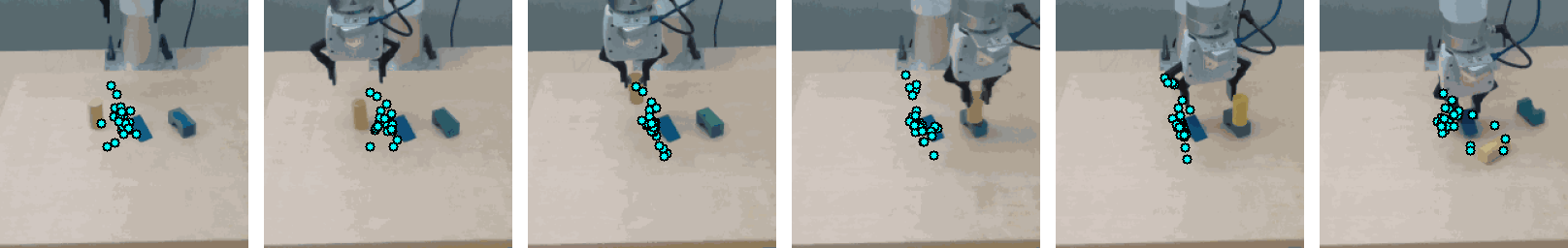}
        \caption{Spatial Softmax}
        \label{fig:giraffe-spatial-softmax}
    \end{subfigure}

    \vspace{1mm}

    \begin{subfigure}{0.6\linewidth}
        \centering
        \includegraphics[width=1.0\linewidth]{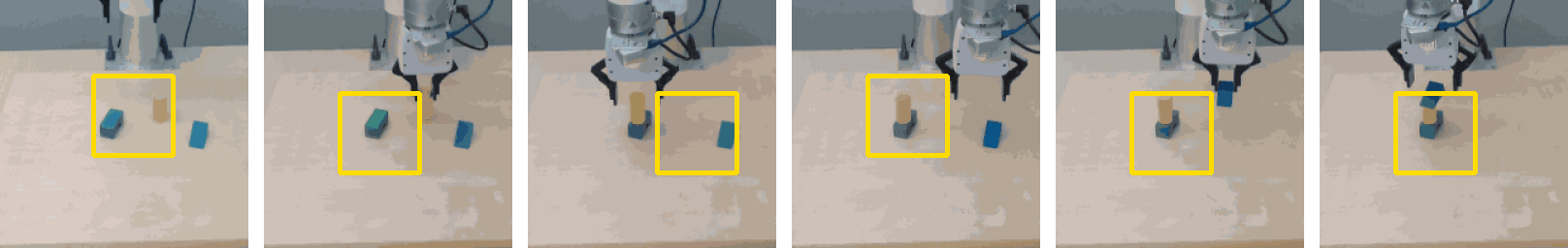}
        \caption{RVT-2}
        \label{fig:giraffe-rvt2}
    \end{subfigure}

    \vspace{1mm}

    \begin{subfigure}{0.6\linewidth}
        \centering
        \includegraphics[width=1.0\linewidth]{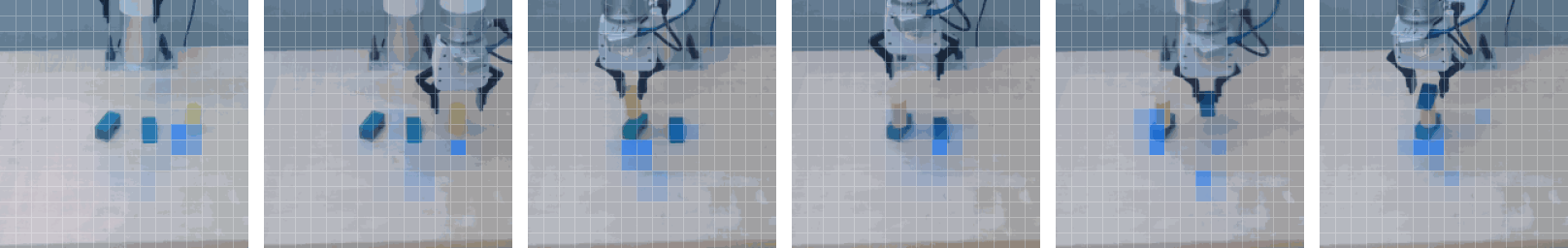}
        \caption{\methodname{}}
        \label{fig:giraffe-focuspool}
    \end{subfigure}
    
    \caption{\textbf{Visual Focus on Giraffe Build.}
    Visualization of Spatial Softmax keypoints, RVT-2 predicted crop regions, and \methodname{} attention-pooling weights overlaid on the input images.}
    \label{fig:real-giraffe-attn}
\end{figure}




\clearpage
\section{Simulation Visualization}

We provide further visualizations of \methodname{} attention-pooling weights and feature activations on six Mimicgen tasks during policy rollouts.

\newcommand{\simfocusfig}[3]{
\begin{figure}[h]
    \centering
    \includegraphics[width=1.0\linewidth]{#1}
    \caption{\textbf{\methodname{} Visualization on #2.}
    Attention-pooling weights and feature activations at residual stage 2 overlaid on input images from one rollout.}
    \label{#3}
\end{figure}
}

\simfocusfig
{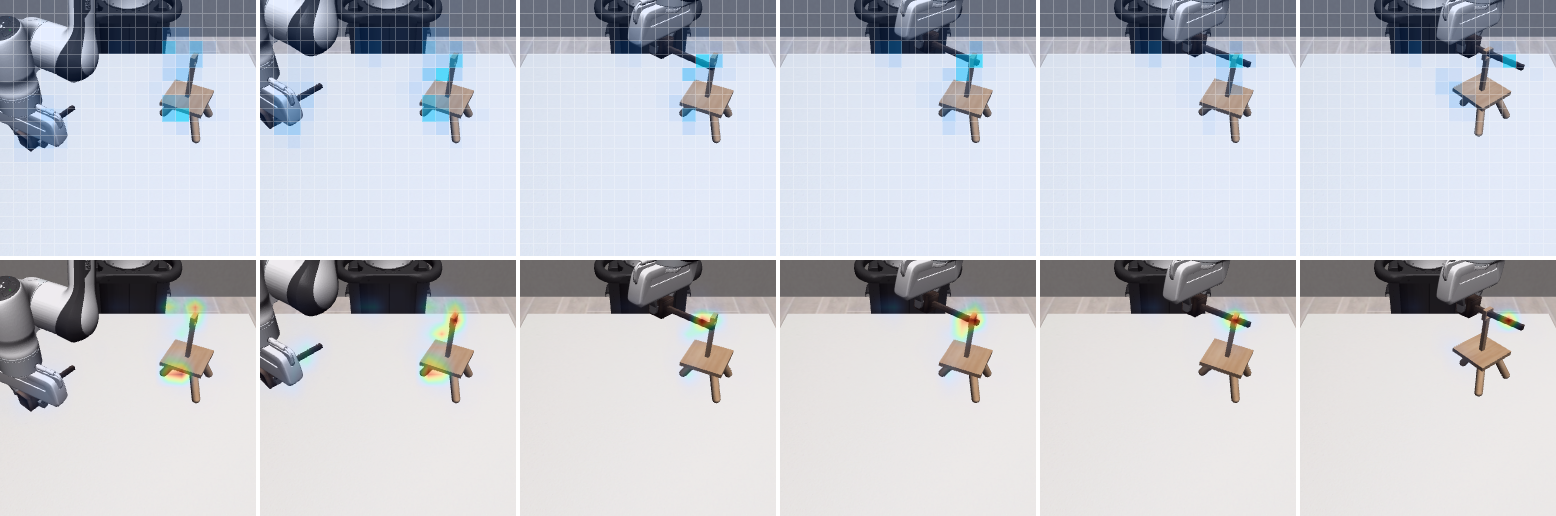}
{Threading D2}
{fig:threading-attn-activation}

\simfocusfig
{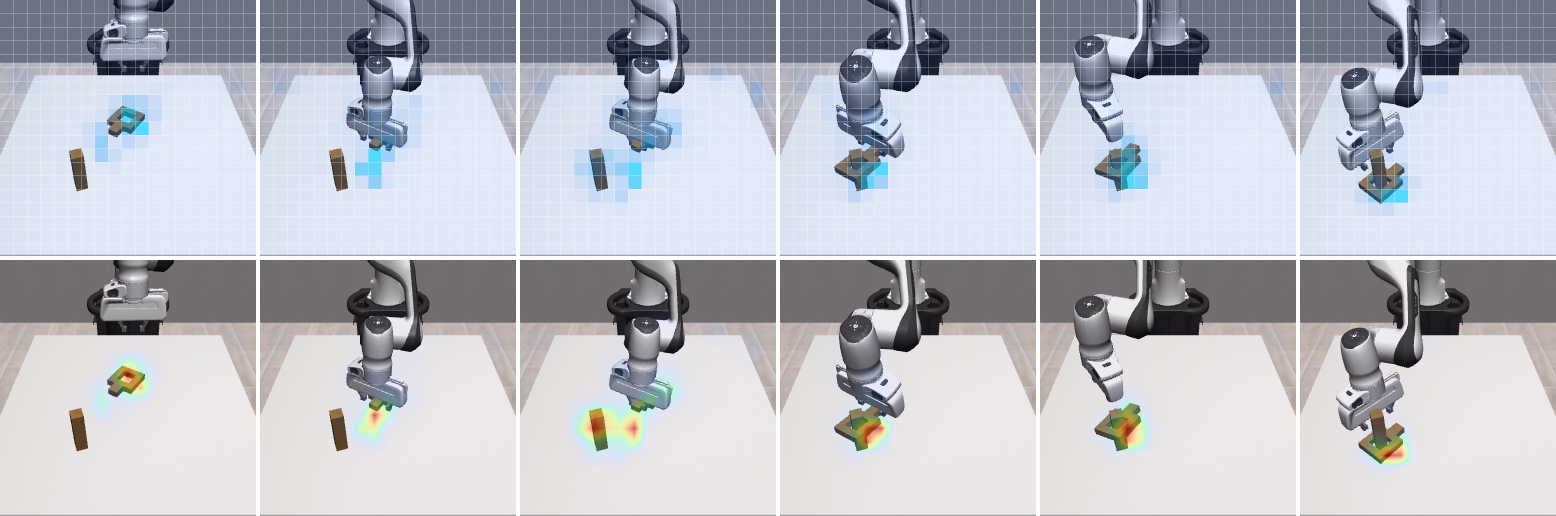}
{Square D2}
{fig:square-attn-activation}

\simfocusfig
{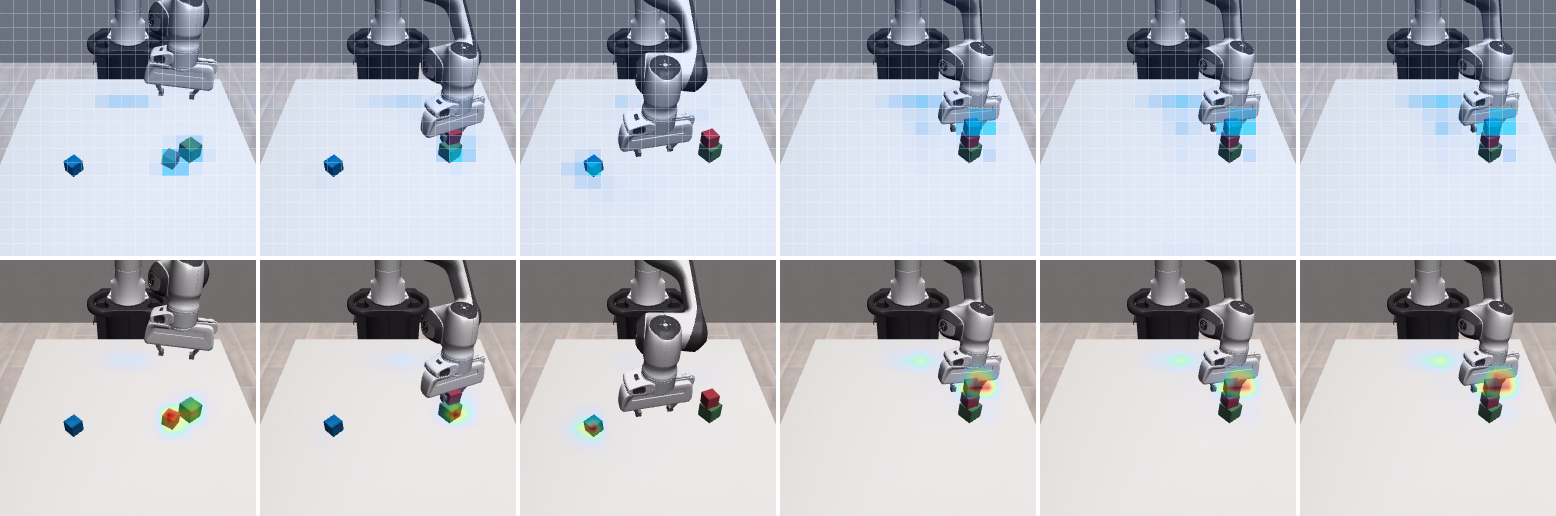}
{Stack Three D1}
{fig:stack-three-attn-activation}

\simfocusfig
{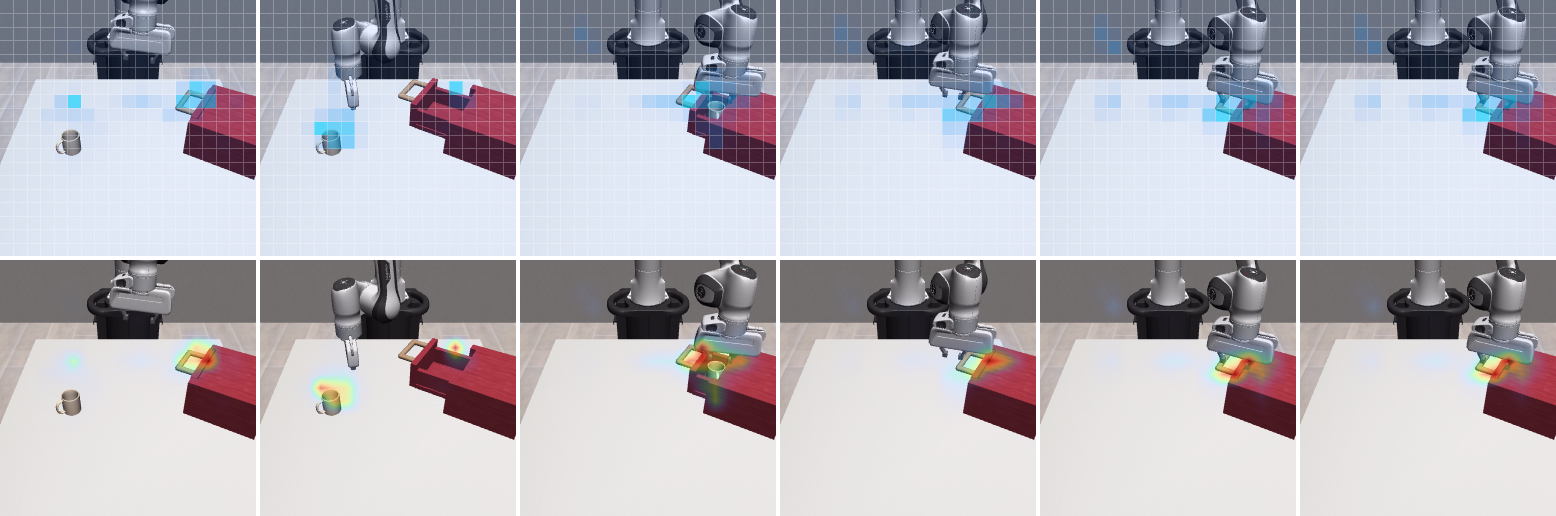}
{Mug Cleanup D1}
{fig:mug-sim-attn-activation}

\simfocusfig
{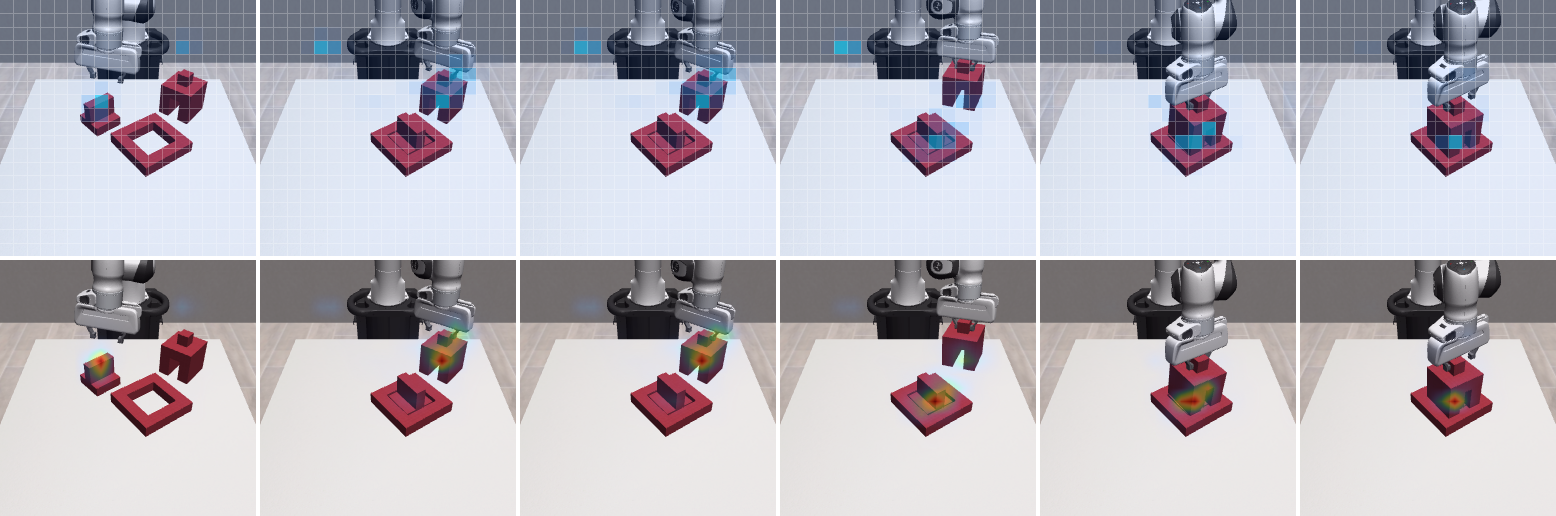}
{Three Piece Assembly D2}
{fig:three-piece-attn-activation}

\simfocusfig
{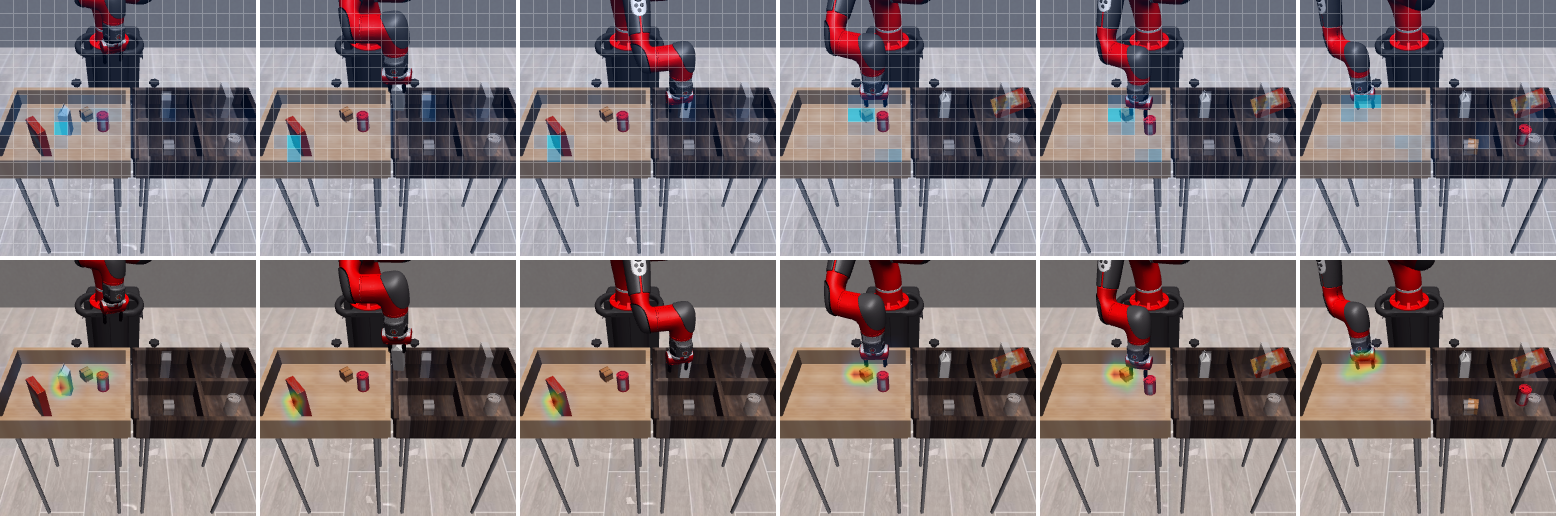}
{Pick and Place D0}
{fig:pnp-sim-attn-activation}
}{}

\end{document}